\documentclass{article} 
\usepackage{iclr_2026_arxiv,times}

\usepackage{amsmath,amsfonts,bm}

\def\eqref#1{equation~\ref{#1}}

\def\1{\bm{1}}

\def\vmu{{\bm{\mu}}}

\def\vd{{\bm{d}}}

\def\vh{{\bm{h}}}

\def\vm{{\bm{m}}}

\def\vo{{\bm{o}}}
\def\vp{{\bm{p}}}
\def\vq{{\bm{q}}}
\def\vr{{\bm{r}}}

\def\vt{{\bm{t}}}

\def\vv{{\bm{v}}}

\def\vx{{\bm{x}}}

\def\mD{{\bm{D}}}

\def\mI{{\bm{I}}}

\def\mR{{\bm{R}}}
\def\mS{{\bm{S}}}

\def\mV{{\bm{V}}}

\def\mSigma{{\bm{\Sigma}}}

\DeclareMathAlphabet{\mathsfit}{\encodingdefault}{\sfdefault}{m}{sl}
\SetMathAlphabet{\mathsfit}{bold}{\encodingdefault}{\sfdefault}{bx}{n}

\usepackage[pagebackref,colorlinks,breaklinks,citecolor=gray]{hyperref}
\usepackage{url}
\usepackage{graphicx}
\usepackage{comment}
\usepackage{amsmath,amssymb,amsfonts,mathabx,mathbbol}
\usepackage{acronym}
\usepackage{enumitem}
\PassOptionsToPackage{dvipsnames,table}{xcolor}
\usepackage{hyperref}
\usepackage{balance}
\usepackage{xspace}
\usepackage{setspace}
\usepackage[skip=3pt,font=small]{subcaption}
\usepackage[skip=3pt,font=small]{caption}
\usepackage{caption}
\usepackage[misc]{ifsym}
\usepackage[capitalise]{cleveref}
\usepackage{tabularx}
\usepackage{multirow,multirow,array,makecell}
\usepackage{overpic}
\usepackage{wrapfig}
\usepackage{pifont}
\usepackage{url}
\usepackage{booktabs,tabularx,colortbl}       
\usepackage[utf8]{inputenc} 
\usepackage[T1]{fontenc}    
\usepackage{nicefrac}       
\usepackage[nopatch=footnote]{microtype}      
\usepackage[noend,ruled, vlined]{algorithm2e}
\usepackage{algorithmic}
\usepackage{rotating}
\usepackage{adjustbox}
\usepackage{hvfloat}
\usepackage{ifsym}
\makeatletter
\DeclareRobustCommand\onedot{\futurelet\@let@token\@onedot}
\def\@onedot{\ifx\@let@token.\else.\null\fi\xspace}

\makeatother

\crefname{algorithm}{Alg.}{Algs.}
\Crefname{algocf}{Algorithm}{Algorithms}
\crefname{section}{Sec.}{Secs.}
\Crefname{section}{Section}{Sections}
\crefname{table}{Tab.}{Tabs.}
\Crefname{table}{Table}{Tables}
\crefname{figure}{Fig.}{Fig.}
\Crefname{figure}{Figure}{Figure}
\definecolor{revision}{RGB}{0,0,255}

\newcommand{\model}{\text{VideoArtGS}\xspace}

\acrodef{3dgs}[3DGS]{3D Gaussian Splatting}
\acrodef{tsdf}[TSDF]{Truncated Signed Distance Function}
\acrodef{fps}[FPS]{Farthest Point Sampling}

\title{\model: Building Digital Twins of Articulated Objects from Monocular Video}
\iclrfinalcopy
\author{
Yu Liu$^{1,2,*}$ \quad Baoxiong Jia$^{2,\dagger,\textrm{ \Letter}}$ \quad Ruijie Lu$^{2,3}$ \quad Chuyue Gan$^{1}$ \quad Huayu Chen$^{1}$ \\
\textbf{Junfeng Ni$^{1,2}$} \quad \textbf{Song-Chun Zhu}$^{1,2,3}$ \quad \textbf{Siyuan Huang}$^{2,\textrm{ \Letter}}$
\vspace{3pt}
\\
\small $^1$Tsinghua University~$^2$State Key Laboratory of General Artificial Intelligence, BIGAI~$^3$\small Peking University\vspace{1pt}\\
}

\begin{document}

\maketitle

\begin{abstract}
Building digital twins of articulated objects from monocular video presents an essential challenge in computer vision, which requires simultaneous reconstruction of object geometry, part segmentation, and articulation parameters from limited viewpoint inputs. Monocular video offers an attractive input format due to its simplicity and scalability; however, it's challenging to disentangle the object geometry and part dynamics with visual supervision alone, as the joint movement of the camera and parts leads to ill-posed estimation. While motion priors from pre-trained tracking models can alleviate the issue, how to effectively integrate them for articulation learning remains largely unexplored.
To address this problem, we introduce \model, a novel approach that reconstructs high-fidelity digital twins of articulated objects from monocular video. We propose a motion prior guidance pipeline that analyzes 3D tracks, filters noise, and provides reliable initialization of articulation parameters. We also design a hybrid center-grid part assignment module for articulation-based deformation fields that captures accurate part motion. 
\model demonstrates state-of-the-art performance in articulation and mesh reconstruction, reducing the reconstruction error by about \textbf{two orders of magnitude} compared to existing methods. \model enables practical digital twin creation from monocular video, establishing a new benchmark for video-based articulated object reconstruction. Our work is made publicly available at: \url{https://videoartgs.github.io}.
\end{abstract}

\let\thefootnote\relax\footnote{$^{*}$Work done as an intern at BIGAI. \quad $^{\dagger}$Project Lead. \quad $\textrm{ \Letter}$ Corresponding authors.}

\section{Introduction}
\label{sec:intro}
Articulated objects, prevalent in our daily life, are becoming a major focus in recent research for computer vision and robotics~\citep{weng2024neural,luo2024physpart,liu2024cage,deng2024articulate,yang2023reconstructing,liu2024survey}.
 Reconstructing interactable digital twins of articulated objects from visual observations is fundamental to advancing applications in augmented reality, robotics simulation, and interactive scene understanding. By generating digital twins from simple inputs like video, we can significantly accelerate the development of intelligent systems, particularly by bridging the sim-to-real gap for robotic manipulation and interaction tasks~\citep{torne2024reconciling,kerr2024rsrd}. 
 To build powerful and generalizable robotic systems, reconstructing interactable objects from monocular video represents a critical frontier, as this would unlock the ability to learn from the vast amount of videos available online and allow robots to model the world through their own eyes.

Recent approaches to reconstructing articulated objects can be broadly categorized into two families based on the way to estimate articulation parameters. One family employs a feed-forward model to predict articulation parameters directly~\citep{mandi2024real2code, le2024articulate, jiang2022ditto}. These methods, however, struggle with scalability and generalization, as they require extensive training on annotated data, which often fails to transfer to novel, real-world settings. Creating datasets that comprehensively cover the sheer combinatorial complexity of real-world objects, articulation types, and viewing conditions is practically infeasible. 
A second, more common family reconstructs objects by explicitly estimating joint parameters from multi-view images of the object in two or more discrete states~\citep{liu2025building, jiayi2023paris, weng2024neural, lin2025splart, yu2025part}.  While these methods benefit from strong geometric constraints, they require controlled, often cumbersome, data capture setups that limit their use outside the lab. This approach is not only constrained by impractical data capture requirements but is also highly brittle; slight misalignments in the coordinate frames between states can cause catastrophic failures in prediction accuracy. 
A far more practical and scalable paradigm is reconstructing articulated objects from casually captured monocular videos, which enables the ability to learn from internet videos and allows robotic agents to model objects directly from their visual observations.

However, the convenience of video capture introduces a profound technical challenge: the reconstruction problem becomes fundamentally ill-posed. From a single, moving viewpoint, the observed pixel motion results from four entangled factors: camera trajectory, object geometry, part segmentation, and articulation-based part dynamics. Disentangling these variables without the strong parallax cues from multi-view data is highly ambiguous. Consequently, prior video-based methods often produce distorted geometries, fail to segment parts correctly, or are confined to overly simplistic objects~\citep{kerr2024rsrd, song2024reacto, peng2025generalizable}, leaving robust, general-purpose reconstruction from monocular video a largely unsolved frontier. 
To break this ambiguity, motion priors from tracking models offer a promising direction. Previous methods, such as Shape-of-Motion~\citep{wang2024shape} and ArtiPoint~\citep{werby2025articulated}, have explored lifting 2D tracks for supervision. More recently, the advent of powerful perception models like SpatialTrackerV2~\citep{xiao2025spatialtracker} and TAPIP3D~\citep{tapip3d} provides 3D tracks, which offer richer motion information. However, both lifted tracks and 3D tracks contain substantial noise that makes them ineffective for direct use in articulated object reconstruction, leaving the problem of how to effectively leverage them as motion priors unexplored.

To address these challenges, we propose \model, which introduces several key innovations for reconstructing articulated objects from monocular video. 
Central to our approach are two key insights: (1) motion priors from pre-trained tracking models are essential for disambiguating object movement, and (2) by enforcing articulation constraints (e.g., linear or circular trajectories for prismatic and revolute joints), we leverage both object-part movement priors and the reconstruction objective to jointly suppress noise in the tracks, recover structural cues of the moving parts.
Specifically, we design a novel motion prior guidance pipeline that analyzes raw 3D tracking trajectories, filters noise, classifies motion types (e.g., revolute, prismatic), and clusters points into coherent parts. This process yields accurate initial estimates for the joint parameters and part centers, transforming the intractable joint optimization into a well-posed refinement problem. To further enhance reconstruction quality, we design a hybrid center-grid part assignment module. This module combines the strengths of spatial clustering for distinct movable parts with a flexible grid-based representation to model complex 3D geometry of objects, enabling clean part segmentation and precise deformation modeling.

These designs enable \model to achieve state-of-the-art performance, reducing reconstruction and articulation estimation errors by approximately two orders of magnitude compared to previous methods on both simple two-part objects and on our new, challenging \model-20 dataset. 
Our approach opens new possibilities for practical digital twin creation from readily available video data, with applications in scenarios where multi-state capture is impractical or impossible. Through extensive experiments, we demonstrate the effectiveness of our method in delivering high-quality reconstruction of articulated objects from monocular video sequences.

\paragraph{Contributions} Our main contributions of this work can be summarized as follows:

\begin{itemize}[leftmargin=*,nolistsep,noitemsep]
\item We propose \model, a novel method for articulated object reconstruction from monocular video that achieves state-of-the-art performance, reducing key error metrics by up to two orders of magnitude over prior work.
\item We introduce a motion prior guidance framework that analyzes 3D tracking trajectories to robustly initialize the deformation field, making the ill-posed reconstruction problem tractable. We design a hybrid center-grid part assignment module that accurately segments parts and benefit articulation learning, accommodating complex geometries.
\item We conduct extensive experiments and establish a new benchmark for video-based articulated object reconstruction, validating the practical applicability of our approach. Our comprehensive ablation studies systematically validate our designs and point out directions for future improvement.
\end{itemize}

\section{Related Work}
\label{sec:related_work}
\subsection{Dynamic Scene Reconstruction}
Dynamic scene reconstruction is a long-standing challenge in computer vision. A significant line of work focuses on jointly estimating camera poses and scene geometry, often represented as depth maps or point clouds. Pioneering methods like DROID-SLAM~\citep{teed2021droid}, CasualSAM~\citep{tang2025causal}, and Mega-SaM~\citep{li2025megasam} established robust frameworks for this task. More recently, foundation models have emerged, with DUSt3R~\citep{wang2024dust3r} and VGGT~\citep{wang2025vggt} providing a powerful basis for 3D reconstruction. Subsequent works like MonST3R~\citep{zhang2024monst3r}, CUT3R~\citep{wang2025continuous}, and SpatialTrackerV2~\citep{xiao2025spatialtracker} have fine-tuned or extended DUSt3R or VGGT to better handle dynamic content. 

While the above methods provide camera and geometry information, representing the dynamic scene itself has been revolutionized by 3D Gaussian Splatting~\citep{kerbl20233d}. Many 4D extensions learn to deform Gaussians implicitly over time~\citep{jung2023deformable, katsumata2023efficient,wu20244d,luiten2024dynamic,li2024spacetime,lu20243d, lei2024gart,guo2024motion,qian20243dgs,bae2024per, liu2025modgs, wu20254d}, which excels at capturing complex, non-rigid motion but offers no explicit control over an object's underlying structure. Although some methods learn dense tracks by reconstructing videos~\citep{som2024, lei2024mosca}, they do not model the articulated object and cannot reconstruct interactive assets from it.
Attempts to add control via superpoints~\citep{huang2024sc} or physics engines~\citep{xie2024physgaussian, jiang2024vr} have been made, but they either fail to extract accurate physical parameters or require impractical priors. \model bridges this gap by integrating an explicit articulation model directly into the deformable Gaussian framework, enabling high-fidelity reconstruction for articulated objects.

\subsection{Articulated Object Reconstruction}
Reconstructing articulated objects presents a dual challenge: one must solve for both the part-level geometry and the underlying articulation parameters. 
One family of methods employs end-to-end models to predict both part segmentation and joint parameters~\citep{heppert2023carto, wei2022self, kawana2021unsupervised, mandi2024real2code, jiang2022ditto, ma2023sim2real, nie2022structure, hsu2023ditto, goyal2025geopard, xia2025drawer}, while some similar methods only predict articulation parameters~\citep{hu2017learning, yi2018deep, li2020category, wang2019shape2motion, sun2023opdmulti, liu2022toward, weng2021captra, sturm2011probabilistic, chu2023command, martin2016integrated, liu2023self, gadre2021act, mo2021where2act, jain2021screwnet, yan2020rpm, lei2023nap}. Their fundamental limitation, however, is a reliance on large, annotated datasets, which prevents them from generalizing to unseen object categories.
The dominant paradigm relies on multi-view observations at discrete multi-state~\citep{liu2025building, tseng2022cla, mu2021sdf, lewis2022narf22, jiayi2023paris, lei2024gart, deng2024articulate, swaminathan2024leia, noguchi2022watch, zhang2021strobenet, pillai2015learning, liu2023building, wang2025self, lewis2025splatart,zhang2025iaao}. These methods leverage strong geometric constraints, which simplify the problem but require impractical and controlled data capture setups. A more practical but far more challenging setting is reconstruction from a monocular video. Existing video-based methods are typically limited to simple objects~\citep{song2024reacto, peng2025generalizable} or rely on a pre-trained segmentation model that has limited generalization ability~\citep{zhou2025monomobility}. In contrast, \model is designed for this challenging setting. By introducing a robust motion prior guidance pipeline, we effectively disentangle the scene dynamics and transform the ill-posed problem into a tractable one, achieving state-of-the-art results where prior methods have struggled.

\section{Method}
\label{sec:method}
\begin{figure}[t!]
 \centering
 \resizebox{\linewidth}{!}{\includegraphics[width=\linewidth]{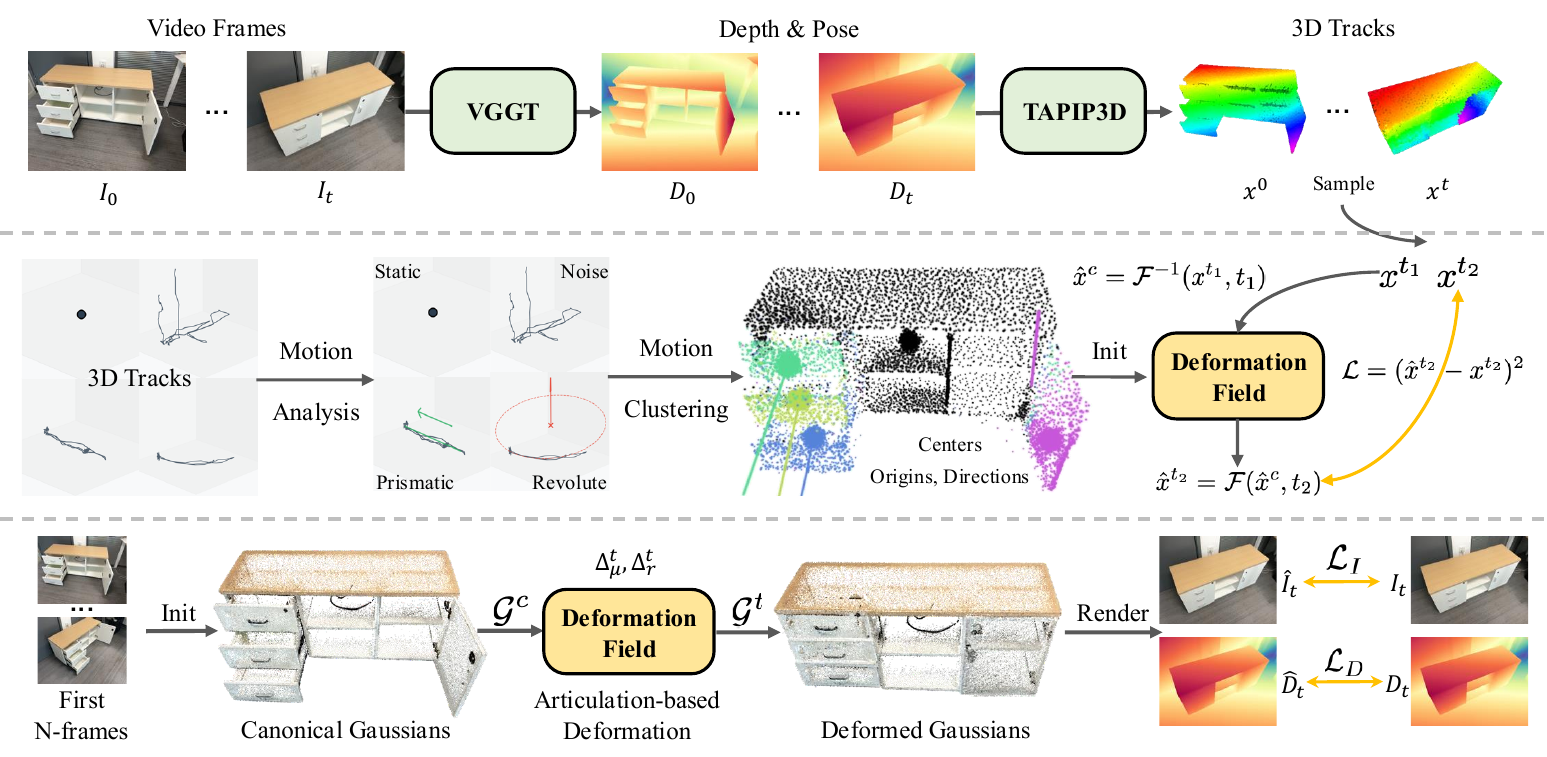}}
 \caption{\textbf{The overview of \model.} Given video frames, we first use VGGT~\citep{wang2025vggt} to estimate the depths along with camera poses and then use TAPIP3D~\citep{tapip3d} to obtain 3D tracks. We design a motion prior guidance pipeline to analyze and group these tracks, initializing our articulation-based deformation field with motion information and optimizing it with tracking loss. Finally, we reconstruct the object with 3D Gaussians and the deformation field, jointly optimizing both modules by rendering and tracking loss.}
 \label{fig:overview}
 \vspace{-10pt}
\end{figure}

Given a monocular video sequence $\{{I}_t\}_{t=1}^T$,  \model reconstructs articulated objects with part meshes $\mathcal{M}$ and articulation parameters $\Psi$. We first use the VGGT~\citep{wang2025vggt} trained for dynamic scenes from SpatialTrackerV2~\citep{xiao2025spatialtracker} to estimate the depths and camera poses, and then reconstruct the object with 3D Gaussians $\mathcal{G} = \{G_i\}_{i=1}^N$ and an articulation-based deformation field $\mathcal{F}$. This field contains a part segmentation module $S_\phi$ and articulation parameters $\Psi$ (including axis directions $\vd$, axis origins $\vo$, and time-variant joint states $\theta_t$) that control the dynamics of each part. We also introduce motion prior from a pre-trained tracking model TAPIP3D~\citep{tapip3d} to guide the initialization and optimization of the deformation field.
An overview of \model is presented in~\cref{fig:overview}, with details on key designs provided in the following sections.

\subsection{Articulation-based Deformation Field}
\label{sec:method:dynamic_modeling}
To model the temporal dynamics of an articulated object, we formulate an articulation-based deformation field $\mathcal{F}$ that transforms a set of canonical Gaussians $G_i^c = \{\vmu_i^c, \vr_i^c, \bm{s}_i, \sigma_i, \vh_i\}$ into the observation state $G_i^t = \{\vmu_i^t, \vr_i^t, \bm{s}_i, \sigma_i, \vh_i\}$ for any given time $t$. Since articulation is a rigid process, the intrinsic properties of each Gaussian—its scale ($\bm{s}_i^c$), opacity ($\sigma_i^c$), and appearance ($\vh_i^c$)—are treated as time-invariant, while its position ($\vmu_i^t$) and rotation ($\vr_i^t$) are time-variant. Following ArtGS~\citep{liu2025building}, \model first assigns each Gaussian to object parts through a segmentation module $S_{\phi}(\cdot)$ and then applies the corresponding rigid transformation for each part:
\begin{equation}
\vm_i = S_\phi(G^c_i), \quad G_i^t = \sum_{k=1}^K m_{ik} \cdot \mathcal{T}_{k}^t(G_i^c)
\end{equation}
where $\vm_i = [m_{i1}, \ldots, m_{iK}]$ represents the assignment probabilities of $i$-th Gaussian to $K$ parts, and $\mathcal{T}_k^t$ denotes the rigid transformation for $k$-th part at time $t$. The number of movable parts and joint types (revolute or prismatic) could be obtained by GPT4-o~\citep{hurst2024gpt}. We provide more details of articulation modeling in \cref{app:imp:arti}.

\paragraph{Hybrid Ce nter-grid Part Assignment}
To effectively assign Gaussians to articulable parts, ArtGS~\citep{liu2025building} proposes a center-based part assignment module that segments parts using the Mahalanobis distance between Gaussians and learnable centers. However, this approach faces limitations when the static base part has complex geometries. A simple but key observation is that static regions remain fixed in space. Unlike movable parts that naturally form distinct motion clusters, the static base part is better characterized by its fixed spatial volume rather than a movable center. We therefore propose a hybrid center-grid part assignment module that combines two strategies.
For the $K-1$ movable parts, we define learnable part centers $C_k = (\vp_k, \mV_k, {\bm \lambda}_k)$ with center location $\vp_k\in\mathbb{R}^3$, rotation matrix $\mV_k\in\mathbb{R}^{3\times 3}$, and scale vector ${\bm \lambda}_k\in\mathbb{R}^3$. For the static base part, we use a learnable hash grid $H$ to model its spatial region directly.
Given the canonical position $\vmu_i^c$ of a Gaussian $G_i^c$, we compute its assignment probabilities $\vm_i$ by fusing scores from both models.
First, we compute the squared Mahalanobis distance $\mD_{i,k}$ to each of the $K-1$ movable part centers:
\begin{equation}
\mD_{i,k} = \left( \frac{\mV_k(\vmu^c_i-\vp_k)}{{\bm\lambda}_k} \right)^\top \left( \frac{\mV_k(\vmu^c_i-\vp_k)}{{\bm\lambda}k} \right) + \Delta_{i, k},
\label{eq:dist}
\end{equation}
where $\Delta_{i,k}$ is a residual term for improving boundary identification that is introduced by ArtGS~\citep{liu2025building}.
Simultaneously, we query the hash grid at the Gaussian's position to get a feature vector, which is processed by a small MLP to produce a single logit, $l_{i} = \text{MLP}(H(\vmu_i^c))$, representing the "staticness" score.
The final assignment probabilities $\vm_i \in \mathbb{R}^K$ are obtained by concatenating the static logit with the negative distances of the movable parts and applying a softmax function:
\begin{equation}
\vm_i = \mathrm{Softmax}\left(\mathrm{concat}\left(\left[l_{i}, -\mD_{i,1}, \dots, -\mD_{i,K-1}\right]\right)\right).
\label{eq:part_assignment}
\end{equation}
This hybrid formulation enables robust segmentation by leveraging both structured geometric relationships for movable parts and flexible spatial modeling for complex static regions.

\subsection{Motion Prior Guidance}
We use a pre-trained tracking model TAPIP3D~\citep{tapip3d} to obtain 3D tracking trajectories, providing a motion prior to guide the initialization and optimization of the deformation field.

\paragraph{Motion Pattern Analysis.}
To identify noises and extract motion information from tracking trajectories, we first analyze the motion pattern of each trajectory and divide all trajectories into 4 classes: static, prismatic, revolute, and noise.
If the maximum displacement distance of the $i$-th trajectory $\{\vx^t_i\}_{t=1}^{T}$ is below a threshold $\epsilon_s$, it is classified as a static trajectory.
For the remaining dynamic trajectories, we use line fitting and circle fitting to identify the motion type and motion parameters.
A main challenge is that all points remain static for most of the time and move for only a short period of time, which is particularly prominent for objects with multiple parts.
Many points are concentrated in the same area, leading to the fitting collapse. 
To deal with this problem, we design an adaptive spatial downsampling approach. Specifically, we first voxelize each trajectory, and then retain only one point in each voxel. To handle different ranges of trajectories, we dynamically adjust the voxel size based on the range of the trajectory.
After downsampling, we use the remaining points to fit the trajectory.

For prismatic motion, we use Principal Component Analysis (PCA) for line fitting, combined with the RANSAC algorithm to improve robustness.
For revolute motion, we first fit the best plane, then fit a circle on that plane. We use Singular Value Decomposition (SVD) to find the normal vector and verify whether the trajectory conforms to rigid rotation.
If the line/circle fitting error of a trajectory is less than pre-defined thresholds $\epsilon_l$/$\epsilon_c$, it is considered a valid track; otherwise, it is treated as a noise track. For a valid track, we prioritize models with smaller fitting errors. 
The above process also provides the direction of prismatic trajectories and the direction and origin of revolute trajectories.

\paragraph{Motion Clustering.}
Given valid trajectories with their motion type and motion parameters, we construct feature vectors and then use K-means clustering to group trajectories into different parts.
For prismatic motion, the feature vector contains starting position, average position, motion direction, and normalized velocity.
For revolute motion, the feature vector contains starting position, average position, axis direction, axis origin, and angular velocity.
To improve clustering quality, we adopt an iterative filtering strategy, combining directional angle filtering and Euclidean distance filtering to remove outliers. 
Finally, we generate articulation information for core parameters initialization of the deformation field $\mathcal{F}$, including the axis direction $\vd$, axis origin $\vo$, and part centers $\vp$.

\paragraph{Deformation Field Initialization.}
We randomly initialize the remaining parameters of $\mathcal{F}$ and then use the tracking trajectories to optimize them. We design two different losses $\mathcal{L}_{c2o}$ and $\mathcal{L}_{o2o}$ to optimize the deformation field. $\mathcal{L}_{c2o}$ is the canonical-to-observation loss, which provides direct supervision for the deformation from canonical state to the observation state:
\begin{equation}
\hat \vx^t_i=\mathcal{F}(\vx^c_i, t),\quad \mathcal{L}_{c2o}=\frac{1}{N}\sum_{i=1}^N(\vx^t_i-\hat \vx^t_i)^2,\\
\label{eq:track_c2o}
\end{equation}
where $\vx_i^c, \vx_i^t$ are point positions sampled from trajectory $\{\vx^t_i\}_{t=1}^{T}$.
$\mathcal{L}_{o2o}$ is the observation-to-observation loss, which enhances temporal consistency between two observation states $t_0$ and $t_1$:
\begin{equation}
\hat \vx^c_i=\mathcal{F}^{-1}(\vx^{t_0}_i, t_0),\quad \hat \vx^{t_1}_i=\mathcal{F}(\hat \vx^{c}_i, t_1), \quad \mathcal{L}_{o2o}=\frac{1}{N}\sum_{i=1}^N(\vx^{t_1}_i-\hat \vx^{t_1}_i)^2, \\
\label{eq:track_o2o}
\end{equation}
where $\mathcal{F}^{-1}$ is the inversed deformation field of $\mathcal{F}$ and $\mathcal{F}^{-1}$ shares the same parameters with $\mathcal{F}$. See \cref{app:inv_df} for details of the inverse deformation field.
We randomly sample pairs of tracking trajectories at time $(c, t)$ for $\mathcal{L}_{c2o}$ and $(t_0, t_1)$ within a 30-frame window for $\mathcal{L}_{o2o}$ to robustly optimize $\mathcal{F}$. The final tracking loss could be calculated by:
\begin{equation}
\begin{aligned}
\mathcal{L}_\text{track}=\mathcal{L}_{c2o}+\mathcal{L}_{o2o}.\\
\end{aligned}
\end{equation}

\subsection{Optimization}
To reconstruct high-quality geometry of objects, we assume the object remains static in the first $N$ frames and initialize canonical Gaussians $\mathcal{G}^c$ with these frames. We train $\mathcal{G}^c$ with the rendering loss $\mathcal{L}_\text{render} = (1-\lambda_{\text{SSIM}})\mathcal{L}_1 + \lambda_{\text{SSIM}}\mathcal{L}_{\text{D-SSIM}}+\mathcal{L}_D$ used in ArtGS~\citep{liu2025building}, where $\mathcal{L}_D = \log\left(1 + ||\mD - \bar{\mD}||_1\right)$ is a depth loss. 
After initializing the deformation field $\mathcal{F}$ and canonical Gaussians $\mathcal{G}^c$, we jointly optimize $\mathcal{F}$ and $\mathcal{G}^c$ across all video frames and tracking trajectories with rendering loss $\mathcal{L}_\text{render}$ and canonical-to-observation tracking loss $\mathcal{L}_{c2o}$:
\begin{equation}
\begin{aligned}
\mathcal{L}=\mathcal{L}_\text{render}+\lambda_{c2o}\mathcal{L}_{c2o}.\\
\end{aligned}
\end{equation}
We provide more implementation and model training details in~\cref{app:imp}.

\section{Experiments}
\label{sec:exp}

\paragraph{Datasets}
We conduct a comprehensive evaluation on two distinct datasets to assess the performance of existing methods on objects with varying articulation complexity.
(1) Video2Articulation-S, a dataset proposed by~\citep{peng2025generalizable}, which serves as our benchmark for simple articulated objects. It consists of 73 test videos across 11 categories of synthetic objects from the PartNet-Mobility dataset~\citep{xiang2020sapien}, where each object has only a single movable part.
(2) \model-20, a newly curated dataset that evaluates the performance on more complex scenarios. It contains 20 videos of complex articulated objects of 10 categories from PartNet-Mobility, featuring more challenging kinematics with 2 to 9 movable parts per object. 

\paragraph{Metrics} Our evaluation protocol includes metrics for both articulation estimation and mesh reconstruction quality.
For articulation estimation, we measure axis direction error ($\deg$), axis position error (cm), and joint state error ($\deg$ for revolute joints, cm for prismatic joints) between the predicted and ground-truth joint parameters. 
For mesh reconstruction, we assess geometric accuracy using the bi-directional Chamfer Distance (CD). This is computed between the reconstructed mesh and the ground-truth mesh, using 10,000 points uniformly sampled from each surface. We report the CD (in cm) for the whole object (CD-w), the static part (CD-s), and the movable parts (CD-m).

\subsection{Results on Simple Articulated Objects}
\label{sec:exp:two-part}
\paragraph{Experimental Setup} For this benchmark, we use the Video2Articulation-S dataset. We perform a quantitative comparison against three state-of-the-art methods: ArticulateAnything~\citep{le2024articulate}, RSRD~\citep{kerr2024rsrd}, and Video2Articulation~\citep{peng2025generalizable}. Following the standard evaluation protocol established by Video2Articulation~\citep{peng2025generalizable}, all metrics are reported as the mean and standard deviation (mean ± std) across all test videos, and articulation estimation metrics are divided into revolute and prismatic. For a fair and direct comparison, our experimental setup utilizes ground-truth depth and camera poses, and the results for all baseline methods are taken directly from Video2Articulation~\citep{peng2025generalizable}. To ensure consistency with our evaluation, we have converted their reported metrics from meters (m) to centimeters (cm) and from radians to degrees. We also retrain and evaluate VideoAticulation on this dataset.

\paragraph{Results and Analysis}

\begin{figure}[t!]
    \centering
    \resizebox{\linewidth}{!}{\includegraphics[width=\linewidth]{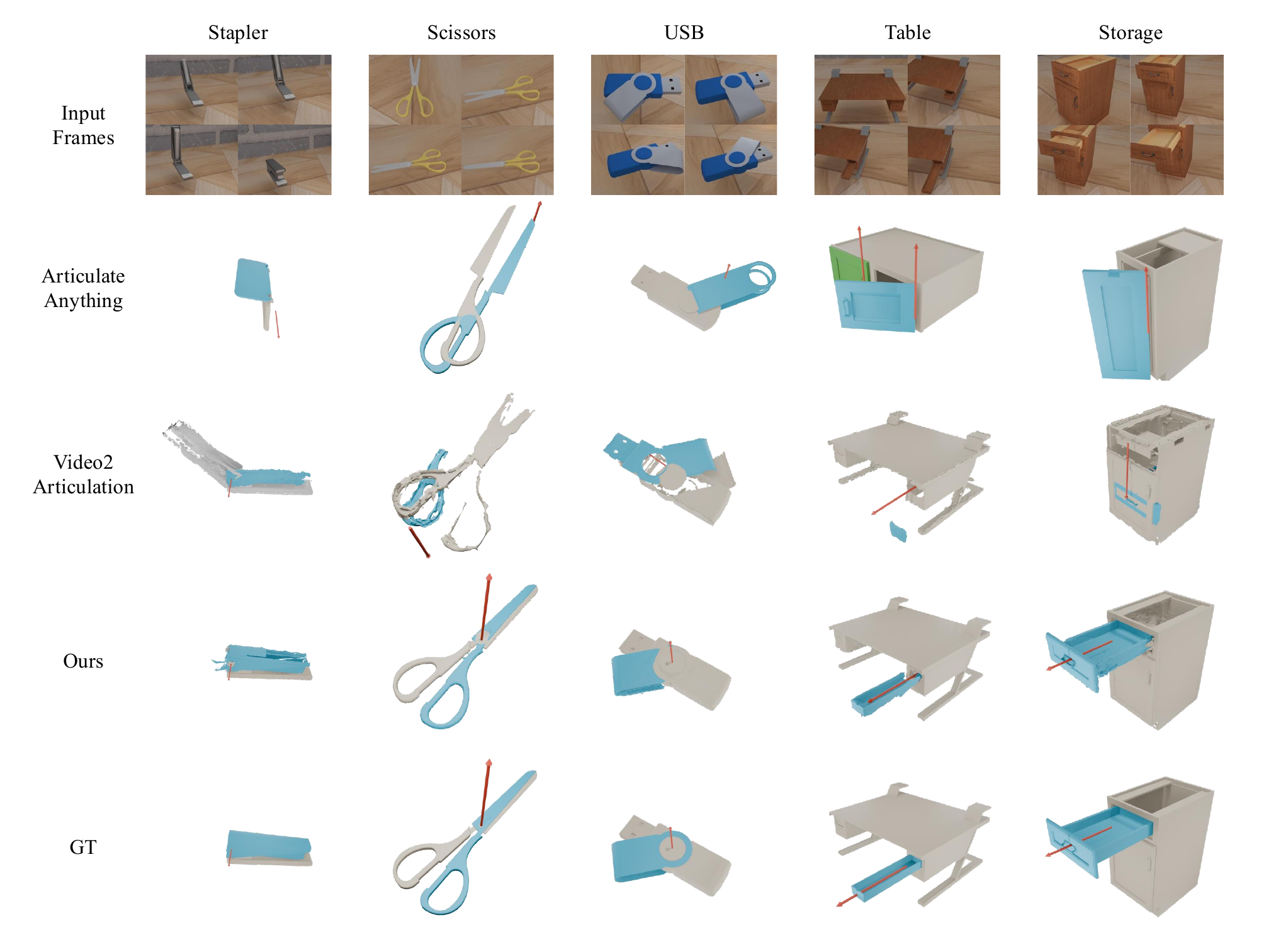}}
    \vspace{-10pt}
    \caption{\textbf{Qualitative results on Video2Articulation-S dataset.} We present reconstruction comparisons between baselines and our model on the Video2Articulation-S dataset. }
    \label{fig:2part}
    \vspace{-10pt}
\end{figure}

The quantitative results, presented in \cref{tab:exp_v2a}, demonstrate that our method substantially outperforms all baseline approaches across all metrics. The most significant gains are in joint parameter estimation, where \model achieves an order-of-magnitude reduction in error compared to the second-best method, Video2Articulation. 
This dramatic increase in accuracy is primarily attributable to our motion prior guidance, which provides an accurate starting point for optimization that prior methods lack.
Our method also achieves a new state of the art in reconstruction quality. The exceptional improvements on both movable parts and the static part validate the effectiveness of \model.
As illustrated in \cref{fig:2part}, our method consistently produces high-fidelity mesh reconstructions with clean part boundaries and precise articulation. This demonstrates the robustness and high quality of our approach across the diverse object categories.

\begin{table*}[t!]
    \caption{\textbf{Quantitative evaluation on Video2Articulation-S dataset.} Metrics are reported as mean $\pm$ std over all test videos. Lower ($\downarrow$) is better on all metrics, and the \textbf{best results} are highlighted in bold. $^\dagger$ means the results are taken from VideoAticulation~\citep{peng2025generalizable}.}
    \label{tab:exp_v2a}
    \renewcommand{\arraystretch}{1.2}
    \resizebox{\linewidth}{!}{
    \begin{tabular}{ccccccccc}
    \toprule
    \multirow{2}{*}{Method} & \multicolumn{3}{c}{Revolute Joint Estimation}                & \multicolumn{2}{c}{Prismatic Joint Estimation} & \multicolumn{3}{c}{Reconstruction}                           \\
    \cmidrule(lr){2-4}\cmidrule(lr){5-6}\cmidrule(lr){7-9}
                        & Axis ($\degree$)   & Position (cm)       & State ($\degree$)  & Axis ($\deg$)          & State (cm)            & CD-w (cm)               & CD-m (cm)               & CD-s (cm)               \\
    \midrule
    ArticulateAnything$^\dagger$~\citep{le2024articulate} & 46.98±45.27     & 81.00±40.00          & N/A             & 52.71±44.69     & N/A         & 11.00±22.00     & 59.00±73.00     & 7.00±18.00      \\
    RSRD$^\dagger$~\citep{kerr2024rsrd}               & 67.06±29.22     & 203.00±748.00        & 59.02±34.38      & 69.91±24.07     & 70.00±48.00        & 339.00±2147.00  & 82.00±117.00    & 14.00±41.00     \\
    Video2Articulation$^\dagger$ ~\citep{peng2025generalizable} & 18.34±32.09     & 13.00±25.00          & 14.32±26.35      & 13.75±18.91     & 8.00±22.00         & 1.00±1.00       & 13.00±26.00     & 6.00±19.00      \\
    Video2Articulation~\citep{peng2025generalizable}  & 13.83±28.15     & 11.55±22.39          & 10.25±21.27      & 14.37±19.08     & 3.44±6.25         & 3.45±16.46       & 12.21±24.44     & 5.39±17.09      \\
    \midrule
    \textbf{Ours}      & \textbf{0.32±0.44}       & \textbf{0.42±0.75}      & \textbf{1.15±2.29}        & \textbf{0.35±0.45}       & \textbf{1.03±2.46}    & \textbf{0.29±0.24} & \textbf{0.40±0.32} & \textbf{1.11±2.11} \\
    \bottomrule
    \end{tabular}
    }
    \vspace{-10pt}
\end{table*}

\subsection{Results on Complex Articulated Objects}
\label{sec:exp:multi-part}

\begin{table*}[t]
    \caption{\textbf{Quantitative evaluation on \model-20 dataset.} Metrics are reported as mean $\pm$ std over all test videos. Lower ($\downarrow$) is better on all metrics, and the \textbf{best results} are highlighted in bold.}
    \label{tab:exp_mpart}
    \renewcommand{\arraystretch}{1.2}
    \resizebox{\linewidth}{!}{
    \begin{tabular}{cccccc}
    \toprule
    Method             & Axis ($^\degree$)      & Position(cm)         & CD-w(cm)             & CD-m(cm)             & CD-s(cm)             \\
    \midrule
    ArticulateAnything~\citep{le2024articulate} & 43.65 ± 44.72        & 15.66 ± 36.20                          & 16.10 ± 37.34        & 17.66 ± 36.74        & 16.04 ±  37.36       \\
    Video2Articulation~\citep{peng2025generalizable}  & 48.88 ± 24.18      & 37.04 ± 31.82      & 5.07 ± 21.78       & 30.63 ± 25.64      & 10.22 ± 22.23      \\
    \midrule
    \textbf{Ours}      & \textbf{0.34±0.80}     & \textbf{0.10±0.10}   & \textbf{0.09±0.09}   & \textbf{0.26±0.61}   & \textbf{0.24±0.58}  \\
    \bottomrule
    \end{tabular}
    }
    \vspace{-10pt}
\end{table*}

\begin{figure}[t!]
    \centering
    \resizebox{\linewidth}{!}{\includegraphics[width=\linewidth]{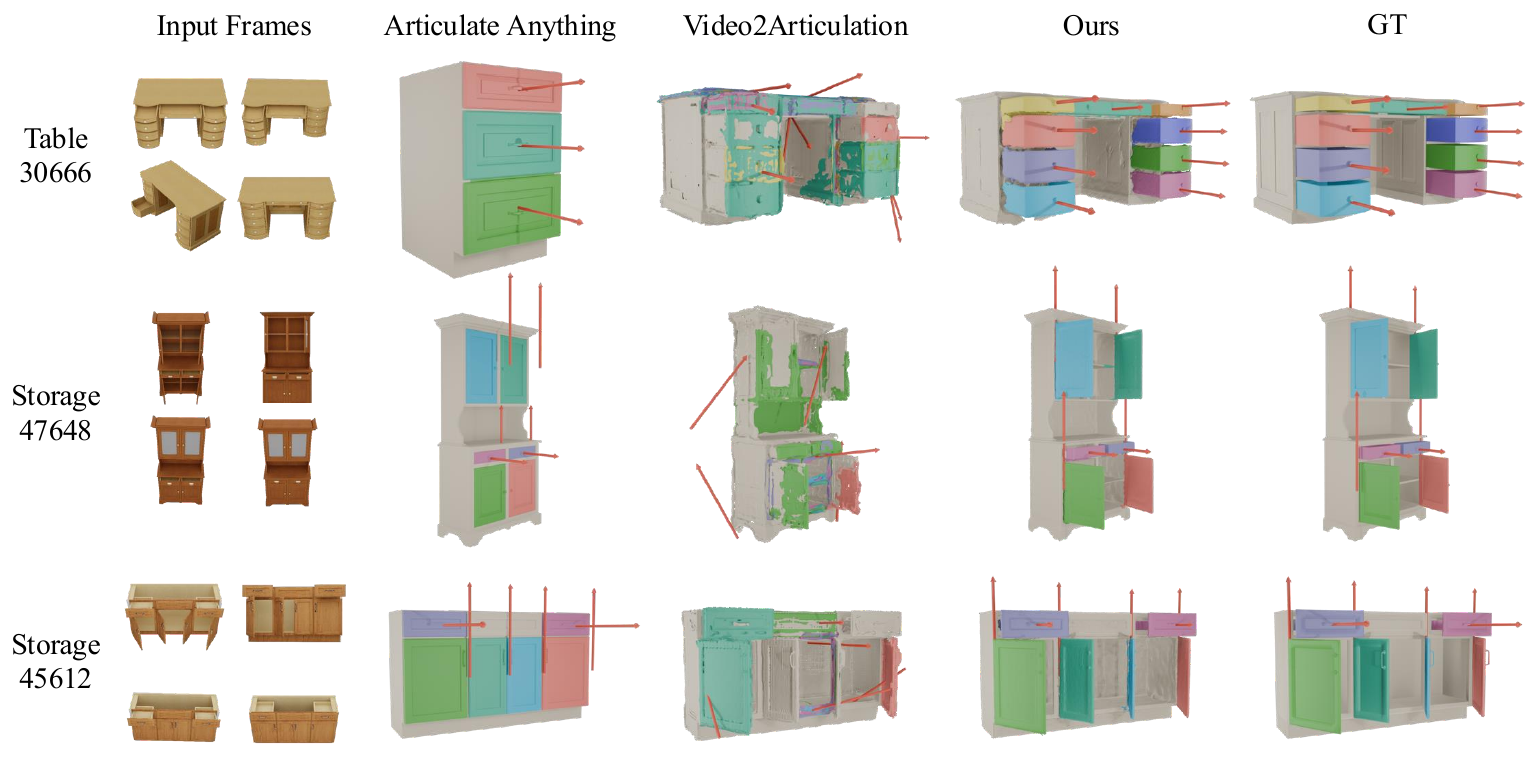}}
    \vspace{-10pt}
    \caption{\textbf{Qualitative results on \model-20 dataset.} We present reconstruction comparisons between baselines and our model on the \model-20 dataset. }
    \label{fig:mpart}
    \vspace{-10pt}
\end{figure}

\paragraph{Experimental Setup}
We conduct an evaluation on our newly curated \model-20 dataset, which contains complex, multi-part objects. We compare our method against current state-of-the-art methods ArticulateAnything~\citep{le2024articulate} and Video2Articulation~\citep{peng2025generalizable}. As RSRD~\citep{kerr2024rsrd} failed to correctly segment parts, we don't use it as a baseline. All metrics are averaged across all parts and reported as mean ± std over all objects.
A critical limitation of prior work is that Video2Articulation~\citep{peng2025generalizable} is designed only for a single movable part. To establish a baseline, we extend it to multi-part objects: we manually isolate video segments where only a single part is in motion and then merge the moving map to extract multiple part meshes. 

\paragraph{Results and Analysis}
On the complex, multi-part \model-20 dataset, our method's advantages become even more pronounced. Compared to Video2Articulation-S, \model-20 has larger camera motion and includes more moving parts, posing a greater challenge to existing baselines. As shown in \cref{fig:mpart}, Video2Articulation struggles to accurately segment moving parts, while ArticulateAnything often retrieves incorrect parts. As demonstrated in \cref{tab:exp_mpart}, \model achieves state-of-the-art performance, drastically outperforming baselines in this complex multi-part setting. It is critical to note that the retrieval database of ArticulateAnything~\citep{le2024articulate} contains the ground-truth meshes and joints from PartNet-Mobility, the same source as our test data. Despite this near-oracle condition for the baseline, our method still reduces articulation estimation errors by nearly two orders of magnitude and excels at mesh reconstruction where baselines fail. These results confirm that \model's advantages generalize from simple to complex scenarios, providing a robust and scalable solution for reconstructing articulated objects from monocular video.

\subsection{Results on Real-World Data}
\label{sec:exp:real}
\paragraph{Experimental Setup}
We also validate the effectiveness of \model on real-world data. We capture monocular videos using a mobile phone camera without LiDAR. We use articulated objects of different categories with different numbers of joints to verify the generalization ability of our method. The input to our model is solely the monocular RGB video.

\begin{figure}[t!]
    \centering
    \resizebox{\linewidth}{!}{\includegraphics[width=\linewidth]{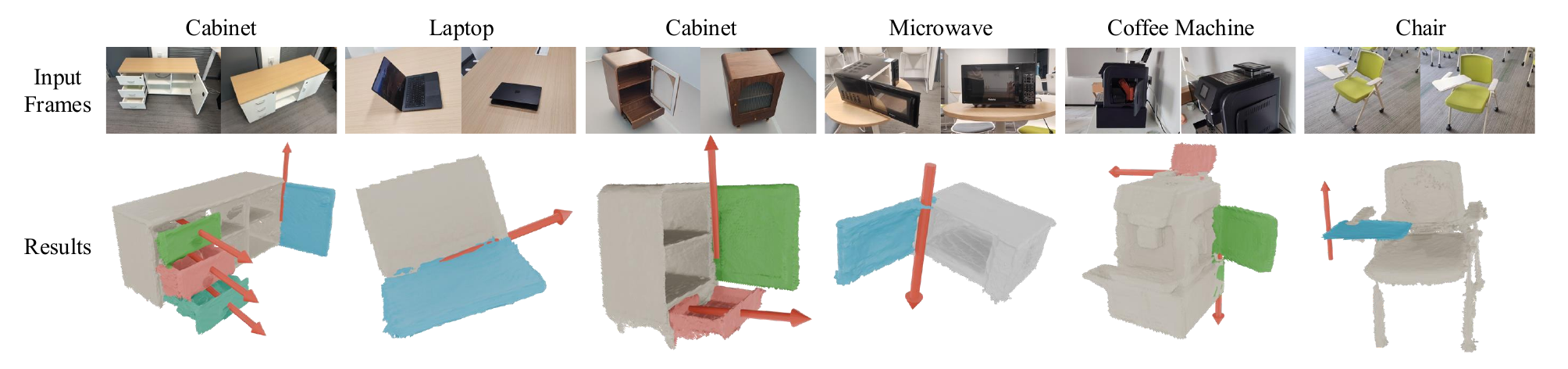}}
    \vspace{-10pt}
    \caption{\textbf{Qualitative results on real-world data.} We present reconstruction results of our model on real-world data, including both simple two-part and complex multi-part objects.}
    \label{fig:real}
    \vspace{-10pt}
\end{figure}

\paragraph{Results and Analysis} As shown in~\cref{fig:real}, our \model successfully reconstructs a diverse set of articulated objects from self-captured, real-world monocular videos, building digital twins with high-fidelity geometry and accurate articulation parameters. \model effectively decouples the object's geometry from its time-varying motion, enabling the creation of a controllable digital asset, fulfilling the promise of creating truly interactable digital twins from casual video captures.

\subsection{Ablation Studies}
\label{sec:exp:ablation}

\paragraph{Experimental Setup} To validate the effectiveness of each component in our method, we conduct comprehensive ablation studies on the \model-20 dataset. We systematically remove different components and analyze their impact on performance, with all metrics reported as mean ± std. 

\begin{table}[t]
    \centering
    \caption{\textbf{Ablation studies on \model-20 dataset.} Lower ($\downarrow$) is better on all metrics, and the \textbf{best results} are highlighted in bold.}
    \label{tab:ablation}
    \renewcommand{\arraystretch}{1.2}
    \resizebox{\linewidth}{!}{
    \begin{tabular}{cccccc}
    \toprule
    \multicolumn{1}{c}{Method}           & Axis ($^\degree$)  & Position (cm)       & CD-w (cm)           & CD-m (cm)           & CD-s (cm)           \\
    \midrule
    Ours                           & \textbf{0.34±0.80} & \textbf{0.10±0.10} & \textbf{0.09±0.09}          & 0.26±0.61          & \textbf{0.24±0.58} \\
    w/o motion prior             & 55.28±15.49        & 23.74±17.49        & 10.18±29.94        & 87.77±17.02        & 14.37±29.69       \\
    w/o center init                & 20.64±21.64        & 22.42±25.03        & 10.33±29.90        & 83.32±14.06        & 14.07±29.80        \\
    w/o deform init & 3.96±3.73          & 2.45±3.07          & 0.11±0.12          & 1.50±2.71          & 0.72±2.05          \\
    w/o axis init                  & 0.60±1.26          & 0.86±2.58          & \textbf{0.09±0.09} & \textbf{0.25±0.56} & 0.27±0.74          \\
    w/o hybrid                     & 1.21±1.77          & 2.51±10.47         & 0.15±0.29          & 10.35±23.84        & 0.50±1.08          \\
    w/o $\mathcal{L}_\text{o2o}$   & 0.68±1.55          & 0.57±1.85          & 0.11±0.12          & 0.58±1.03          & 0.27±0.65          \\
    w/o $\mathcal{L}_\text{c2o}$   & 0.40±0.79          & 0.13±0.11          & 0.09±0.10          & 0.35±0.86          & 0.26±0.70          \\
    \bottomrule
    \end{tabular}
    }
    \vspace{-12pt}
\end{table}

\paragraph{Results and Analysis}
The results, summarized in \cref{tab:ablation}, systematically deconstruct our model's performance and validate the critical role of our core design choices.
\begin{itemize}[leftmargin=*,nolistsep,noitemsep]
\item\textit{Motion Prior Guidance.} The most profound impact comes from removing the entire motion prior guidance (w/o motion prior), including the initialization of centers, joint axes, and deformation field, which results in a catastrophic failure of the model. This unequivocally confirms our central hypothesis: without a strong initial estimate derived from motion cues, the optimization problem of complex articulated objects is intractable.

\item\textit{Initialization of components.}
Removing the part centers initialization (w/o center init) leads to a complete failure, underscoring the necessity of establishing a correct spatial anchor for each part before optimizing its motion. Removing the deformation field initialization (w/o deform init) causes a notable but not catastrophic performance drop, particularly on movable part reconstruction. Interestingly, removing the axis initialization (w/o axis init) yields a marginal drop in articulation estimation and has a minimal effect on reconstruction. This suggests that the framework is robust enough to find the correct axis if the part centers and correspondences are well-initialized, though direct initialization remains beneficial for stability and performance.

\item\textit{Hybrid Center-Grid Assignment.} Replacing the hybrid center-grid assignment module with the center-based assignment module (w/o hybrid) leads to moderate performance drops, especially for the reconstruction of movable parts and articulation estimation. This result highlights that our hybrid assignment is essential for correctly segmenting parts and learning articulation dynamics.

\item\textit{Tracking Losses.} Disabling the observation-to-observation tracking loss (w/o $\mathcal{L}_{o2o}$) degrades performance more than disabling the direct canonical-to-observation loss (w/o $\mathcal{L}_{c2o}$). This indicates that enforcing temporal consistency directly on the observation space is a more critical constraint for achieving precise and stable joint estimation.
\end{itemize}

These ablation results confirm that our method's success relies on the synergistic combination of all components.  The results unequivocally demonstrate that the motion prior guidance and the hybrid part assignment are the two foundational pillars enabling our method's success. The remaining components, while having a smaller individual impact, contribute synergistically to the stability and precision of the final result, solidifying the robustness of our overall framework.

\section{Conclusion}
\label{sec:conclusion}
In conclusion, we introduce \model, a novel method that reconstructs high-fidelity articulated objects from a monocular video. We solve the fundamentally ill-posed challenge by introducing a motion prior guidance pipeline, leveraging 3D tracks to provide robust initialization and optimization of the deformation field. Combined with a hybrid center-grid assignment module for accurate part segmentation, \model achieves a new state of the art, reducing key error metrics by up to two orders of magnitude and validating on our new, challenging \model-20 benchmark.
While \model sets a new performance benchmark, its reliance on upstream trackers, pose estimators, and the necessity of visible motion in the video present avenues for future work. Promising directions include developing end-to-end models that jointly learn tracking and reconstruction or integrating physical priors to handle more challenging, motion-scarce scenarios.

\bibliography{ref}

\begin{thebibliography}{87}
\providecommand{\natexlab}[1]{#1}
\providecommand{\url}[1]{\texttt{#1}}
\expandafter\ifx\csname urlstyle\endcsname\relax
  \providecommand{\doi}[1]{doi: #1}\else
  \providecommand{\doi}{doi: \begingroup \urlstyle{rm}\Url}\fi

\bibitem[Bae et~al.(2024)Bae, Kim, Yun, Lee, Bang, and Uh]{bae2024per}
Jeongmin Bae, Seoha Kim, Youngsik Yun, Hahyun Lee, Gun Bang, and Youngjung Uh.
\newblock Per-gaussian embedding-based deformation for deformable 3d gaussian splatting.
\newblock \emph{arXiv preprint arXiv:2404.03613}, 2024.

\bibitem[Chu et~al.(2023)Chu, Liu, Ye, Tan, Qi, Fu, and Jia]{chu2023command}
Ruihang Chu, Zhengzhe Liu, Xiaoqing Ye, Xiao Tan, Xiaojuan Qi, Chi-Wing Fu, and Jiaya Jia.
\newblock Command-driven articulated object understanding and manipulation.
\newblock In \emph{Proceedings of Conference on Computer Vision and Pattern Recognition (CVPR)}, 2023.

\bibitem[Deng et~al.(2024)Deng, Subr, and Bilen]{deng2024articulate}
Jianning Deng, Kartic Subr, and Hakan Bilen.
\newblock Articulate your nerf: Unsupervised articulated object modeling via conditional view synthesis.
\newblock \emph{arXiv preprint arXiv:2406.16623}, 2024.

\bibitem[Gadre et~al.(2021)Gadre, Ehsani, and Song]{gadre2021act}
Samir~Yitzhak Gadre, Kiana Ehsani, and Shuran Song.
\newblock Act the part: Learning interaction strategies for articulated object part discovery.
\newblock In \emph{Proceedings of International Conference on Computer Vision (ICCV)}, 2021.

\bibitem[Goyal et~al.(2025)Goyal, Petrov, Andrews, Ben-Shabat, Liu, and Kalogerakis]{goyal2025geopard}
Pradyumn Goyal, Dmitry Petrov, Sheldon Andrews, Yizhak Ben-Shabat, Hsueh-Ti~Derek Liu, and Evangelos Kalogerakis.
\newblock Geopard: Geometric pretraining for articulation prediction in 3d shapes.
\newblock \emph{arXiv preprint arXiv:2504.02747}, 2025.

\bibitem[Guo et~al.(2024)Guo, Zhou, Li, Wang, and Li]{guo2024motion}
Zhiyang Guo, Wengang Zhou, Li~Li, Min Wang, and Houqiang Li.
\newblock Motion-aware 3d gaussian splatting for efficient dynamic scene reconstruction.
\newblock \emph{arXiv preprint arXiv:2403.11447}, 2024.

\bibitem[Heppert et~al.(2023)Heppert, Irshad, Zakharov, Liu, Ambrus, Bohg, Valada, and Kollar]{heppert2023carto}
Nick Heppert, Muhammad~Zubair Irshad, Sergey Zakharov, Katherine Liu, Rares~Andrei Ambrus, Jeannette Bohg, Abhinav Valada, and Thomas Kollar.
\newblock Carto: Category and joint agnostic reconstruction of articulated objects.
\newblock In \emph{Proceedings of Conference on Computer Vision and Pattern Recognition (CVPR)}, 2023.

\bibitem[Hsu et~al.(2023)Hsu, Jiang, and Zhu]{hsu2023ditto}
Cheng-Chun Hsu, Zhenyu Jiang, and Yuke Zhu.
\newblock Ditto in the house: Building articulation models of indoor scenes through interactive perception.
\newblock In \emph{Proceedings of International Conference on Robotics and Automation (ICRA)}, 2023.

\bibitem[Hu et~al.(2017)Hu, Li, Van~Kaick, Shamir, Zhang, and Huang]{hu2017learning}
Ruizhen Hu, Wenchao Li, Oliver Van~Kaick, Ariel Shamir, Hao Zhang, and Hui Huang.
\newblock Learning to predict part mobility from a single static snapshot.
\newblock \emph{ACM Transactions on Graphics (TOG)}, 36\penalty0 (6):\penalty0 1--13, 2017.

\bibitem[Huang et~al.(2024)Huang, Sun, Yang, Lyu, Cao, and Qi]{huang2024sc}
Yi-Hua Huang, Yang-Tian Sun, Ziyi Yang, Xiaoyang Lyu, Yan-Pei Cao, and Xiaojuan Qi.
\newblock Sc-gs: Sparse-controlled gaussian splatting for editable dynamic scenes.
\newblock In \emph{Proceedings of Conference on Computer Vision and Pattern Recognition (CVPR)}, 2024.

\bibitem[Hurst et~al.(2024)Hurst, Lerer, Goucher, Perelman, Ramesh, Clark, Ostrow, Welihinda, Hayes, Radford, et~al.]{hurst2024gpt}
Aaron Hurst, Adam Lerer, Adam~P Goucher, Adam Perelman, Aditya Ramesh, Aidan Clark, AJ~Ostrow, Akila Welihinda, Alan Hayes, Alec Radford, et~al.
\newblock Gpt-4o system card.
\newblock \emph{arXiv preprint arXiv:2410.21276}, 2024.

\bibitem[Jain et~al.(2021)Jain, Lioutikov, Chuck, and Niekum]{jain2021screwnet}
Ajinkya Jain, Rudolf Lioutikov, Caleb Chuck, and Scott Niekum.
\newblock Screwnet: Category-independent articulation model estimation from depth images using screw theory.
\newblock In \emph{Proceedings of International Conference on Robotics and Automation (ICRA)}, 2021.

\bibitem[Jiang et~al.(2024)Jiang, Yu, Xie, Li, Feng, Wang, Li, Lau, Gao, Yang, et~al.]{jiang2024vr}
Ying Jiang, Chang Yu, Tianyi Xie, Xuan Li, Yutao Feng, Huamin Wang, Minchen Li, Henry Lau, Feng Gao, Yin Yang, et~al.
\newblock Vr-gs: a physical dynamics-aware interactive gaussian splatting system in virtual reality.
\newblock In \emph{ACM SIGGRAPH 2024 Conference Papers}, 2024.

\bibitem[Jiang et~al.(2022)Jiang, Hsu, and Zhu]{jiang2022ditto}
Zhenyu Jiang, Cheng-Chun Hsu, and Yuke Zhu.
\newblock Ditto: Building digital twins of articulated objects from interaction.
\newblock In \emph{Proceedings of Conference on Computer Vision and Pattern Recognition (CVPR)}, 2022.

\bibitem[Jung et~al.(2023)Jung, Brasch, Song, Perez-Pellitero, Zhou, Li, Navab, and Busam]{jung2023deformable}
HyunJun Jung, Nikolas Brasch, Jifei Song, Eduardo Perez-Pellitero, Yiren Zhou, Zhihao Li, Nassir Navab, and Benjamin Busam.
\newblock Deformable 3d gaussian splatting for animatable human avatars.
\newblock \emph{arXiv preprint arXiv:2312.15059}, 2023.

\bibitem[Katsumata et~al.(2023)Katsumata, Vo, and Nakayama]{katsumata2023efficient}
Kai Katsumata, Duc~Minh Vo, and Hideki Nakayama.
\newblock An efficient 3d gaussian representation for monocular/multi-view dynamic scenes.
\newblock \emph{arXiv preprint arXiv:2311.12897}, 2023.

\bibitem[Kawana et~al.(2021)Kawana, Mukuta, and Harada]{kawana2021unsupervised}
Yuki Kawana, Yusuke Mukuta, and Tatsuya Harada.
\newblock Unsupervised pose-aware part decomposition for 3d articulated objects.
\newblock \emph{arXiv preprint arXiv:2110.04411}, 2021.

\bibitem[Kerbl et~al.(2023)Kerbl, Kopanas, Leimk{\"u}hler, and Drettakis]{kerbl20233d}
Bernhard Kerbl, Georgios Kopanas, Thomas Leimk{\"u}hler, and George Drettakis.
\newblock 3d gaussian splatting for real-time radiance field rendering.
\newblock \emph{ACM Trans. Graph.}, 42\penalty0 (4):\penalty0 139--1, 2023.

\bibitem[Kerr et~al.(2024)Kerr, Kim, Wu, Yi, Wang, Goldberg, and Kanazawa]{kerr2024rsrd}
Justin Kerr, Chung~Min Kim, Mingxuan Wu, Brent Yi, Qianqian Wang, Ken Goldberg, and Angjoo Kanazawa.
\newblock Robot see robot do: Imitating articulated object manipulation with monocular 4d reconstruction.
\newblock In \emph{Conference on Robot Learning (CoRL)}, 2024.

\bibitem[Kirillov et~al.(2023)Kirillov, Mintun, Ravi, Mao, Rolland, Gustafson, Xiao, Whitehead, Berg, Lo, et~al.]{kirillov2023segment}
Alexander Kirillov, Eric Mintun, Nikhila Ravi, Hanzi Mao, Chloe Rolland, Laura Gustafson, Tete Xiao, Spencer Whitehead, Alexander~C Berg, Wan-Yen Lo, et~al.
\newblock Segment anything.
\newblock In \emph{Proceedings of International Conference on Computer Vision (ICCV)}, 2023.

\bibitem[Le et~al.(2025)Le, Xie, Liang, Wang, Yang, Ma, Vedder, Krishna, Jayaraman, and Eaton]{le2024articulate}
Long Le, Jason Xie, William Liang, Hung-Ju Wang, Yue Yang, Yecheng~Jason Ma, Kyle Vedder, Arjun Krishna, Dinesh Jayaraman, and Eric Eaton.
\newblock Articulate-anything: Automatic modeling of articulated objects via a vision-language foundation model.
\newblock In \emph{Proceedings of International Conference on Learning Representations (ICLR)}, 2025.

\bibitem[Lei et~al.(2023)Lei, Deng, Shen, Guibas, and Daniilidis]{lei2023nap}
Jiahui Lei, Congyue Deng, William~B Shen, Leonidas~J Guibas, and Kostas Daniilidis.
\newblock Nap: Neural 3d articulated object prior.
\newblock 2023.

\bibitem[Lei et~al.(2024{\natexlab{a}})Lei, Wang, Pavlakos, Liu, and Daniilidis]{lei2024gart}
Jiahui Lei, Yufu Wang, Georgios Pavlakos, Lingjie Liu, and Kostas Daniilidis.
\newblock Gart: Gaussian articulated template models.
\newblock In \emph{Proceedings of Conference on Computer Vision and Pattern Recognition (CVPR)}, 2024{\natexlab{a}}.

\bibitem[Lei et~al.(2024{\natexlab{b}})Lei, Weng, Harley, Guibas, and Daniilidis]{lei2024mosca}
Jiahui Lei, Yijia Weng, Adam Harley, Leonidas Guibas, and Kostas Daniilidis.
\newblock {MoSca}: Dynamic gaussian fusion from casual videos via {4D} motion scaffolds.
\newblock \emph{arXiv preprint arXiv:2405.17421}, 2024{\natexlab{b}}.

\bibitem[Lewis et~al.(2022)Lewis, Pavlasek, and Jenkins]{lewis2022narf22}
Stanley Lewis, Jana Pavlasek, and Odest~Chadwicke Jenkins.
\newblock Narf22: Neural articulated radiance fields for configuration-aware rendering.
\newblock In \emph{Proceedings of International Conference on Intelligent Robots and Systems (IROS)}, 2022.

\bibitem[Lewis et~al.(2025)Lewis, Chandra, Gao, and Jenkins]{lewis2025splatart}
Stanley Lewis, Vishal Chandra, Tom Gao, and Odest~Chadwicke Jenkins.
\newblock Splatart: Articulated gaussian splatting with estimated object structure.
\newblock \emph{arXiv preprint arXiv:2506.12184}, 2025.

\bibitem[Li et~al.(2020)Li, Wang, Yi, Guibas, Abbott, and Song]{li2020category}
Xiaolong Li, He~Wang, Li~Yi, Leonidas~J Guibas, A~Lynn Abbott, and Shuran Song.
\newblock Category-level articulated object pose estimation.
\newblock In \emph{Proceedings of Conference on Computer Vision and Pattern Recognition (CVPR)}, 2020.

\bibitem[Li et~al.(2024)Li, Chen, Li, and Xu]{li2024spacetime}
Zhan Li, Zhang Chen, Zhong Li, and Yi~Xu.
\newblock Spacetime gaussian feature splatting for real-time dynamic view synthesis.
\newblock In \emph{Proceedings of Conference on Computer Vision and Pattern Recognition (CVPR)}, 2024.

\bibitem[Li et~al.(2025)Li, Tucker, Cole, Wang, Jin, Ye, Kanazawa, Holynski, and Snavely]{li2025megasam}
Zhengqi Li, Richard Tucker, Forrester Cole, Qianqian Wang, Linyi Jin, Vickie Ye, Angjoo Kanazawa, Aleksander Holynski, and Noah Snavely.
\newblock Megasam: Accurate, fast and robust structure and motion from casual dynamic videos.
\newblock In \emph{CVPR}, 2025.

\bibitem[Lin et~al.(2025)Lin, Fang, Irshad, Guizilini, Ambrus, Shakhnarovich, and Walter]{lin2025splart}
Shengjie Lin, Jiading Fang, Muhammad~Zubair Irshad, Vitor~Campagnolo Guizilini, Rares~Andrei Ambrus, Greg Shakhnarovich, and Matthew~R Walter.
\newblock Splart: Articulation estimation and part-level reconstruction with 3d gaussian splatting.
\newblock \emph{arXiv preprint arXiv:2506.03594}, 2025.

\bibitem[Liu et~al.(2023{\natexlab{a}})Liu, Mahdavi-Amiri, and Savva]{jiayi2023paris}
Jiayi Liu, Ali Mahdavi-Amiri, and Manolis Savva.
\newblock Paris: Part-level reconstruction and motion analysis for articulated objects.
\newblock In \emph{Proceedings of International Conference on Computer Vision (ICCV)}, 2023{\natexlab{a}}.

\bibitem[Liu et~al.(2024{\natexlab{a}})Liu, Savva, and Mahdavi-Amiri]{liu2024survey}
Jiayi Liu, Manolis Savva, and Ali Mahdavi-Amiri.
\newblock Survey on modeling of articulated objects.
\newblock \emph{arXiv preprint arXiv:2403.14937}, 2024{\natexlab{a}}.

\bibitem[Liu et~al.(2024{\natexlab{b}})Liu, Tam, Mahdavi-Amiri, and Savva]{liu2024cage}
Jiayi Liu, Hou In~Ivan Tam, Ali Mahdavi-Amiri, and Manolis Savva.
\newblock Cage: Controllable articulation generation.
\newblock In \emph{Proceedings of Conference on Computer Vision and Pattern Recognition (CVPR)}, 2024{\natexlab{b}}.

\bibitem[Liu et~al.(2025)Liu, Iliash, Chang, Savva, and Mahdavi-Amiri]{liu2024singapo}
Jiayi Liu, Denys Iliash, Angel~X Chang, Manolis Savva, and Ali Mahdavi-Amiri.
\newblock Singapo: Single image controlled generation of articulated parts in objects.
\newblock In \emph{Proceedings of International Conference on Learning Representations (ICLR)}, 2025.

\bibitem[Liu et~al.(2022)Liu, Xue, Xu, Fu, and Lu]{liu2022toward}
Liu Liu, Han Xue, Wenqiang Xu, Haoyuan Fu, and Cewu Lu.
\newblock Toward real-world category-level articulation pose estimation.
\newblock \emph{Proceedings of Transactions on Image Processing (TIP)}, 31:\penalty0 1072--1083, 2022.

\bibitem[LIU et~al.(2025)LIU, Liu, Wang, Lyu, Wang, Wang, and Hou]{liu2025modgs}
Qingming LIU, Yuan Liu, Jiepeng Wang, Xianqiang Lyu, Peng Wang, Wenping Wang, and Junhui Hou.
\newblock Mo{DGS}: Dynamic gaussian splatting from casually-captured monocular videos with depth priors.
\newblock In \emph{The Thirteenth International Conference on Learning Representations}, 2025.
\newblock URL \url{https://openreview.net/forum?id=2prShxdLkX}.

\bibitem[Liu et~al.(2023{\natexlab{b}})Liu, Gupta, and Wang]{liu2023building}
Shaowei Liu, Saurabh Gupta, and Shenlong Wang.
\newblock Building rearticulable models for arbitrary 3d objects from 4d point clouds.
\newblock In \emph{Proceedings of Conference on Computer Vision and Pattern Recognition (CVPR)}, 2023{\natexlab{b}}.

\bibitem[Liu et~al.(2023{\natexlab{c}})Liu, Zhang, Hu, Huang, Wang, and Yi]{liu2023self}
Xueyi Liu, Ji~Zhang, Ruizhen Hu, Haibin Huang, He~Wang, and Li~Yi.
\newblock Self-supervised category-level articulated object pose estimation with part-level se (3) equivariance.
\newblock In \emph{Proceedings of International Conference on Learning Representations (ICLR)}, 2023{\natexlab{c}}.

\bibitem[Liu et~al.(2025)Liu, Jia, Lu, Ni, Zhu, and Huang]{liu2025building}
Yu~Liu, Baoxiong Jia, Ruijie Lu, Junfeng Ni, Song-Chun Zhu, and Siyuan Huang.
\newblock Building interactable replicas of complex articulated objects via gaussian splatting.
\newblock In \emph{Proceedings of International Conference on Learning Representations (ICLR)}, 2025.

\bibitem[Lu et~al.(2024)Lu, Guo, Hui, Chen, Yang, Tang, Zhu, and Dai]{lu20243d}
Zhicheng Lu, Xiang Guo, Le~Hui, Tianrui Chen, Min Yang, Xiao Tang, Feng Zhu, and Yuchao Dai.
\newblock 3d geometry-aware deformable gaussian splatting for dynamic view synthesis.
\newblock In \emph{Proceedings of Conference on Computer Vision and Pattern Recognition (CVPR)}, 2024.

\bibitem[Luiten et~al.(2024)Luiten, Kopanas, Leibe, and Ramanan]{luiten2024dynamic}
Jonathon Luiten, Georgios Kopanas, Bastian Leibe, and Deva Ramanan.
\newblock Dynamic 3d gaussians: Tracking by persistent dynamic view synthesis.
\newblock In \emph{Proceedings of International Conference on 3D Vision (3DV)}, 2024.

\bibitem[Luo et~al.(2025)Luo, Geng, Deng, Li, Wang, Jia, Guibas, and Huang]{luo2024physpart}
Rundong Luo, Haoran Geng, Congyue Deng, Puhao Li, Zan Wang, Baoxiong Jia, Leonidas Guibas, and Siyuang Huang.
\newblock Physpart: Physically plausible part completion for interactable objects.
\newblock In \emph{Proceedings of International Conference on Robotics and Automation (ICRA)}, 2025.

\bibitem[Ma et~al.(2023)Ma, Meng, Liu, Chen, Xu, and Chen]{ma2023sim2real}
Liqian Ma, Jiaojiao Meng, Shuntao Liu, Weihang Chen, Jing Xu, and Rui Chen.
\newblock Sim2real 2: Actively building explicit physics model for precise articulated object manipulation.
\newblock In \emph{Proceedings of International Conference on Robotics and Automation (ICRA)}, 2023.

\bibitem[Mandi et~al.(2024)Mandi, Weng, Bauer, and Song]{mandi2024real2code}
Zhao Mandi, Yijia Weng, Dominik Bauer, and Shuran Song.
\newblock Real2code: Reconstruct articulated objects via code generation.
\newblock \emph{arXiv preprint arXiv:2406.08474}, 2024.

\bibitem[Mart{\'\i}n-Mart{\'\i}n et~al.(2016)Mart{\'\i}n-Mart{\'\i}n, H{\"o}fer, and Brock]{martin2016integrated}
Roberto Mart{\'\i}n-Mart{\'\i}n, Sebastian H{\"o}fer, and Oliver Brock.
\newblock An integrated approach to visual perception of articulated objects.
\newblock In \emph{Proceedings of International Conference on Robotics and Automation (ICRA)}, 2016.

\bibitem[Mo et~al.(2021)Mo, Guibas, Mukadam, Gupta, and Tulsiani]{mo2021where2act}
Kaichun Mo, Leonidas~J Guibas, Mustafa Mukadam, Abhinav Gupta, and Shubham Tulsiani.
\newblock Where2act: From pixels to actions for articulated 3d objects.
\newblock In \emph{Proceedings of International Conference on Computer Vision (ICCV)}, 2021.

\bibitem[Mu et~al.(2021)Mu, Qiu, Kortylewski, Yuille, Vasconcelos, and Wang]{mu2021sdf}
Jiteng Mu, Weichao Qiu, Adam Kortylewski, Alan Yuille, Nuno Vasconcelos, and Xiaolong Wang.
\newblock A-sdf: Learning disentangled signed distance functions for articulated shape representation.
\newblock In \emph{Proceedings of International Conference on Computer Vision (ICCV)}, 2021.

\bibitem[Nie et~al.(2022)Nie, Gadre, Ehsani, and Song]{nie2022structure}
Neil Nie, Samir~Yitzhak Gadre, Kiana Ehsani, and Shuran Song.
\newblock Structure from action: Learning interactions for articulated object 3d structure discovery.
\newblock \emph{arXiv preprint arXiv:2207.08997}, 2022.

\bibitem[Noguchi et~al.(2022)Noguchi, Iqbal, Tremblay, Harada, and Gallo]{noguchi2022watch}
Atsuhiro Noguchi, Umar Iqbal, Jonathan Tremblay, Tatsuya Harada, and Orazio Gallo.
\newblock Watch it move: Unsupervised discovery of 3d joints for re-posing of articulated objects.
\newblock In \emph{Proceedings of Conference on Computer Vision and Pattern Recognition (CVPR)}, 2022.

\bibitem[Oquab et~al.(2023)Oquab, Darcet, Moutakanni, Vo, Szafraniec, Khalidov, Fernandez, Haziza, Massa, El-Nouby, et~al.]{oquab2023dinov2}
Maxime Oquab, Timoth{\'e}e Darcet, Th{\'e}o Moutakanni, Huy Vo, Marc Szafraniec, Vasil Khalidov, Pierre Fernandez, Daniel Haziza, Francisco Massa, Alaaeldin El-Nouby, et~al.
\newblock Dinov2: Learning robust visual features without supervision.
\newblock \emph{arXiv preprint arXiv:2304.07193}, 2023.

\bibitem[Peng et~al.(2025)Peng, Lv, Lu, and Savva]{peng2025generalizable}
Weikun Peng, Jun Lv, Cewu Lu, and Manolis Savva.
\newblock Generalizable articulated object reconstruction from casually captured rgbd videos.
\newblock \emph{arXiv preprint arXiv:2506.08334}, 2025.

\bibitem[Pillai et~al.(2015)Pillai, Walter, and Teller]{pillai2015learning}
Sudeep Pillai, Matthew~R Walter, and Seth Teller.
\newblock Learning articulated motions from visual demonstration.
\newblock \emph{arXiv preprint arXiv:1502.01659}, 2015.

\bibitem[Qian et~al.(2024)Qian, Wang, Mihajlovic, Geiger, and Tang]{qian20243dgs}
Zhiyin Qian, Shaofei Wang, Marko Mihajlovic, Andreas Geiger, and Siyu Tang.
\newblock 3dgs-avatar: Animatable avatars via deformable 3d gaussian splatting.
\newblock In \emph{Proceedings of Conference on Computer Vision and Pattern Recognition (CVPR)}, 2024.

\bibitem[Song et~al.(2024)Song, Wei, Foo, Lin, and Liu]{song2024reacto}
Chaoyue Song, Jiacheng Wei, Chuan~Sheng Foo, Guosheng Lin, and Fayao Liu.
\newblock Reacto: Reconstructing articulated objects from a single video.
\newblock In \emph{Proceedings of Conference on Computer Vision and Pattern Recognition (CVPR)}, 2024.

\bibitem[Sturm et~al.(2011)Sturm, Stachniss, and Burgard]{sturm2011probabilistic}
J{\"u}rgen Sturm, Cyrill Stachniss, and Wolfram Burgard.
\newblock A probabilistic framework for learning kinematic models of articulated objects.
\newblock \emph{Journal of Artificial Intelligence Research}, 41, 2011.

\bibitem[Sun et~al.(2023)Sun, Jiang, Savva, and Chang]{sun2023opdmulti}
Xiaohao Sun, Hanxiao Jiang, Manolis Savva, and Angel~Xuan Chang.
\newblock Opdmulti: Openable part detection for multiple objects.
\newblock \emph{arXiv preprint arXiv:2303.14087}, 2023.

\bibitem[Swaminathan et~al.(2024)Swaminathan, Gupta, Gupta, Maiya, Agarwal, and Shrivastava]{swaminathan2024leia}
Archana Swaminathan, Anubhav Gupta, Kamal Gupta, Shishira~R Maiya, Vatsal Agarwal, and Abhinav Shrivastava.
\newblock Leia: Latent view-invariant embeddings for implicit 3d articulation.
\newblock \emph{arXiv preprint arXiv:2409.06703}, 2024.

\bibitem[Tang et~al.(2025)Tang, Xu, Wu, and Lu]{tang2025causal}
Tao Tang, Shijie Xu, Yiting Wu, and Zhixiang Lu.
\newblock Causal-sam-llm: Large language models as causal reasoners for robust medical segmentation.
\newblock \emph{arXiv preprint arXiv:2507.03585}, 2025.

\bibitem[Teed \& Deng(2021)Teed and Deng]{teed2021droid}
Zachary Teed and Jia Deng.
\newblock {DROID-SLAM: Deep Visual SLAM for Monocular, Stereo, and RGB-D Cameras}.
\newblock \emph{Advances in neural information processing systems}, 2021.

\bibitem[Torne et~al.(2024)Torne, Simeonov, Li, Chan, Chen, Gupta, and Agrawal]{torne2024reconciling}
Marcel Torne, Anthony Simeonov, Zechu Li, April Chan, Tao Chen, Abhishek Gupta, and Pulkit Agrawal.
\newblock Reconciling reality through simulation: A real-to-sim-to-real approach for robust manipulation.
\newblock \emph{arXiv preprint arXiv:2403.03949}, 2024.

\bibitem[Tseng et~al.(2022)Tseng, Liao, Yen-Chen, and Sun]{tseng2022cla}
Wei-Cheng Tseng, Hung-Ju Liao, Lin Yen-Chen, and Min Sun.
\newblock Cla-nerf: Category-level articulated neural radiance field.
\newblock In \emph{Proceedings of International Conference on Robotics and Automation (ICRA)}, 2022.

\bibitem[Wang et~al.(2025{\natexlab{a}})Wang, Yuan, Jin, Zhao, Che, Xue, Tian, Huang, and Tang]{wang2025self}
Haowen Wang, Xiaoping Yuan, Zhao Jin, Zhen Zhao, Zhengping Che, Yousong Xue, Jin Tian, Yakun Huang, and Jian Tang.
\newblock Self-supervised multi-part articulated objects modeling via deformable gaussian splatting and progressive primitive segmentation.
\newblock \emph{arXiv preprint arXiv:2506.09663}, 2025{\natexlab{a}}.

\bibitem[Wang et~al.(2025{\natexlab{b}})Wang, Chen, Karaev, Vedaldi, Rupprecht, and Novotny]{wang2025vggt}
Jianyuan Wang, Minghao Chen, Nikita Karaev, Andrea Vedaldi, Christian Rupprecht, and David Novotny.
\newblock Vggt: Visual geometry grounded transformer.
\newblock In \emph{CVPR}, 2025{\natexlab{b}}.

\bibitem[Wang et~al.(2024{\natexlab{a}})Wang, Ye, Gao, Austin, Li, and Kanazawa]{wang2024shape}
Qianqian Wang, Vickie Ye, Hang Gao, Jake Austin, Zhengqi Li, and Angjoo Kanazawa.
\newblock Shape of motion: 4d reconstruction from a single video.
\newblock \emph{arXiv preprint arXiv:2407.13764}, 2024{\natexlab{a}}.

\bibitem[Wang et~al.(2025{\natexlab{c}})Wang, Ye, Gao, Zeng, Austin, Li, and Kanazawa]{som2024}
Qianqian Wang, Vickie Ye, Hang Gao, Weijia Zeng, Jake Austin, Zhengqi Li, and Angjoo Kanazawa.
\newblock Shape of motion: 4d reconstruction from a single video.
\newblock In \emph{International Conference on Computer Vision (ICCV)}, 2025{\natexlab{c}}.

\bibitem[Wang et~al.(2025{\natexlab{d}})Wang, Zhang, Holynski, Efros, and Kanazawa]{wang2025continuous}
Qianqian Wang, Yifei Zhang, Aleksander Holynski, Alexei~A Efros, and Angjoo Kanazawa.
\newblock Continuous 3d perception model with persistent state.
\newblock In \emph{CVPR}, 2025{\natexlab{d}}.

\bibitem[Wang et~al.(2024{\natexlab{b}})Wang, Leroy, Cabon, Chidlovskii, and Revaud]{wang2024dust3r}
Shuzhe Wang, Vincent Leroy, Yohann Cabon, Boris Chidlovskii, and Jerome Revaud.
\newblock Dust3r: Geometric 3d vision made easy.
\newblock In \emph{Proceedings of Conference on Computer Vision and Pattern Recognition (CVPR)}, 2024{\natexlab{b}}.

\bibitem[Wang et~al.(2019)Wang, Zhou, Shi, Chen, Zhao, and Xu]{wang2019shape2motion}
Xiaogang Wang, Bin Zhou, Yahao Shi, Xiaowu Chen, Qinping Zhao, and Kai Xu.
\newblock Shape2motion: Joint analysis of motion parts and attributes from 3d shapes.
\newblock In \emph{Proceedings of Conference on Computer Vision and Pattern Recognition (CVPR)}, 2019.

\bibitem[Wei et~al.(2022)Wei, Chabra, Ma, Lassner, Zollh{\"o}fer, Rusinkiewicz, Sweeney, Newcombe, and Slavcheva]{wei2022self}
Fangyin Wei, Rohan Chabra, Lingni Ma, Christoph Lassner, Michael Zollh{\"o}fer, Szymon Rusinkiewicz, Chris Sweeney, Richard Newcombe, and Mira Slavcheva.
\newblock Self-supervised neural articulated shape and appearance models.
\newblock In \emph{Proceedings of Conference on Computer Vision and Pattern Recognition (CVPR)}, 2022.

\bibitem[Weng et~al.(2021)Weng, Wang, Zhou, Qin, Duan, Fan, Chen, Su, and Guibas]{weng2021captra}
Yijia Weng, He~Wang, Qiang Zhou, Yuzhe Qin, Yueqi Duan, Qingnan Fan, Baoquan Chen, Hao Su, and Leonidas~J Guibas.
\newblock Captra: Category-level pose tracking for rigid and articulated objects from point clouds.
\newblock In \emph{Proceedings of International Conference on Computer Vision (ICCV)}, 2021.

\bibitem[Weng et~al.(2024)Weng, Wen, Tremblay, Blukis, Fox, Guibas, and Birchfield]{weng2024neural}
Yijia Weng, Bowen Wen, Jonathan Tremblay, Valts Blukis, Dieter Fox, Leonidas Guibas, and Stan Birchfield.
\newblock Neural implicit representation for building digital twins of unknown articulated objects.
\newblock In \emph{Proceedings of Conference on Computer Vision and Pattern Recognition (CVPR)}, 2024.

\bibitem[Werby et~al.(2025)Werby, B{\"u}chner, R{\"o}fer, Huang, Burgard, and Valada]{werby2025articulated}
Abdelrhman Werby, Martin B{\"u}chner, Adrian R{\"o}fer, Chenguang Huang, Wolfram Burgard, and Abhinav Valada.
\newblock Articulated object estimation in the wild.
\newblock In \emph{Conference on Robot Learning (CoRL)}, 2025.

\bibitem[Wu et~al.(2025)Wu, Liu, Hung, Qian, Zhan, and Duan]{wu20254d}
Diankun Wu, Fangfu Liu, Yi-Hsin Hung, Yue Qian, Xiaohang Zhan, and Yueqi Duan.
\newblock 4d-fly: Fast 4d reconstruction from a single monocular video.
\newblock In \emph{Proceedings of the Computer Vision and Pattern Recognition Conference}, 2025.

\bibitem[Wu et~al.(2024)Wu, Yi, Fang, Xie, Zhang, Wei, Liu, Tian, and Wang]{wu20244d}
Guanjun Wu, Taoran Yi, Jiemin Fang, Lingxi Xie, Xiaopeng Zhang, Wei Wei, Wenyu Liu, Qi~Tian, and Xinggang Wang.
\newblock 4d gaussian splatting for real-time dynamic scene rendering.
\newblock In \emph{Proceedings of Conference on Computer Vision and Pattern Recognition (CVPR)}, 2024.

\bibitem[Xia et~al.(2024)Xia, Su, Memmel, Jain, Yu, Mbiziwo-Tiapo, Farhadi, Gupta, Wang, and Ma]{xia2025drawer}
Hongchi Xia, Entong Su, Marius Memmel, Arhan Jain, Raymond Yu, Numfor Mbiziwo-Tiapo, Ali Farhadi, Abhishek Gupta, Shenlong Wang, and Wei-Chiu Ma.
\newblock Drawer: Digital reconstruction and articulation with environment realism.
\newblock In \emph{Proceedings of Conference on Computer Vision and Pattern Recognition (CVPR)}, 2024.

\bibitem[Xiang et~al.(2020)Xiang, Qin, Mo, Xia, Zhu, Liu, Liu, Jiang, Yuan, Wang, et~al.]{xiang2020sapien}
Fanbo Xiang, Yuzhe Qin, Kaichun Mo, Yikuan Xia, Hao Zhu, Fangchen Liu, Minghua Liu, Hanxiao Jiang, Yifu Yuan, He~Wang, et~al.
\newblock Sapien: A simulated part-based interactive environment.
\newblock In \emph{Proceedings of Conference on Computer Vision and Pattern Recognition (CVPR)}, 2020.

\bibitem[Xiao et~al.(2025)Xiao, Wang, Xue, Karaev, Makarov, Kang, Zhu, Bao, Shen, and Zhou]{xiao2025spatialtracker}
Yuxi Xiao, Jianyuan Wang, Nan Xue, Nikita Karaev, Iurii Makarov, Bingyi Kang, Xin Zhu, Hujun Bao, Yujun Shen, and Xiaowei Zhou.
\newblock Spatialtrackerv2: 3d point tracking made easy.
\newblock In \emph{ICCV}, 2025.

\bibitem[Xie et~al.(2024)Xie, Zong, Qiu, Li, Feng, Yang, and Jiang]{xie2024physgaussian}
Tianyi Xie, Zeshun Zong, Yuxing Qiu, Xuan Li, Yutao Feng, Yin Yang, and Chenfanfu Jiang.
\newblock Physgaussian: Physics-integrated 3d gaussians for generative dynamics.
\newblock In \emph{Proceedings of Conference on Computer Vision and Pattern Recognition (CVPR)}, 2024.

\bibitem[Yan et~al.(2020)Yan, Hu, Yan, Chen, Van~Kaick, Zhang, and Huang]{yan2020rpm}
Zihao Yan, Ruizhen Hu, Xingguang Yan, Luanmin Chen, Oliver Van~Kaick, Hao Zhang, and Hui Huang.
\newblock Rpm-net: recurrent prediction of motion and parts from point cloud.
\newblock \emph{arXiv preprint arXiv:2006.14865}, 2020.

\bibitem[Yang et~al.(2023)Yang, Wang, Reddy, and Ramanan]{yang2023reconstructing}
Gengshan Yang, Chaoyang Wang, N~Dinesh Reddy, and Deva Ramanan.
\newblock Reconstructing animatable categories from videos.
\newblock In \emph{Proceedings of Conference on Computer Vision and Pattern Recognition (CVPR)}, 2023.

\bibitem[Yi et~al.(2018)Yi, Huang, Liu, Kalogerakis, Su, and Guibas]{yi2018deep}
Li~Yi, Haibin Huang, Difan Liu, Evangelos Kalogerakis, Hao Su, and Leonidas Guibas.
\newblock Deep part induction from articulated object pairs.
\newblock \emph{arXiv preprint arXiv:1809.07417}, 2018.

\bibitem[Yu et~al.(2025)Yu, Shah, Wahed, Shen, Nguyen, and Lourentzou]{yu2025part}
Tianjiao Yu, Vedant Shah, Muntasir Wahed, Ying Shen, Kiet~A Nguyen, and Ismini Lourentzou.
\newblock Part$^2$gs: Part-aware modeling of articulated objects using 3d gaussian splatting.
\newblock \emph{arXiv preprint arXiv:2506.17212}, 2025.

\bibitem[Zhang et~al.(2025)Zhang, Ke, Harley, and Fragkiadaki]{tapip3d}
Bowei Zhang, Lei Ke, Adam~W Harley, and Katerina Fragkiadaki.
\newblock Tapip3d: Tracking any point in persistent 3d geometry.
\newblock \emph{arXiv preprint arXiv:2504.14717}, 2025.

\bibitem[Zhang \& Lee(2025)Zhang and Lee]{zhang2025iaao}
Can Zhang and Gim~Hee Lee.
\newblock Iaao: Interactive affordance learning for articulated objects in 3d environments.
\newblock In \emph{Proceedings of the Computer Vision and Pattern Recognition Conference}, pp.\  12132--12142, 2025.

\bibitem[Zhang et~al.(2021)Zhang, Litany, Sridhar, and Guibas]{zhang2021strobenet}
Ge~Zhang, Or~Litany, Srinath Sridhar, and Leonidas Guibas.
\newblock Strobenet: Category-level multiview reconstruction of articulated objects.
\newblock \emph{arXiv preprint arXiv:2105.08016}, 2021.

\bibitem[Zhang et~al.(2024)Zhang, Herrmann, Hur, Jampani, Darrell, Cole, Sun, and Yang]{zhang2024monst3r}
Junyi Zhang, Charles Herrmann, Junhwa Hur, Varun Jampani, Trevor Darrell, Forrester Cole, Deqing Sun, and Ming-Hsuan Yang.
\newblock Monst3r: A simple approach for estimating geometry in the presence of motion.
\newblock \emph{arXiv preprint arxiv:2410.03825}, 2024.

\bibitem[Zhou et~al.(2025)Zhou, Wang, Guo, and Xu]{zhou2025monomobility}
Hongyi Zhou, Xiaogang Wang, Yulan Guo, and Kai Xu.
\newblock Monomobility: Zero-shot 3d mobility analysis from monocular videos.
\newblock \emph{arXiv preprint arXiv:2505.11868}, 2025.

\end{thebibliography}
\bibliographystyle{iclr2026_conference}

\clearpage

\appendix
\renewcommand{\thefigure}{A.\arabic{figure}}
\renewcommand{\thetable}{A.\arabic{table}}
\renewcommand{\theequation}{A.\arabic{equation}}
\setcounter{section}{0}
\setcounter{figure}{0}
\setcounter{table}{0}
\setcounter{equation}{0}

\section{Implementation and Training Details}
\label{app:imp}
\subsection{\model-20 Dataset} 

\begin{figure}[t!]
    \centering
    \resizebox{\linewidth}{!}{\includegraphics[width=\linewidth]{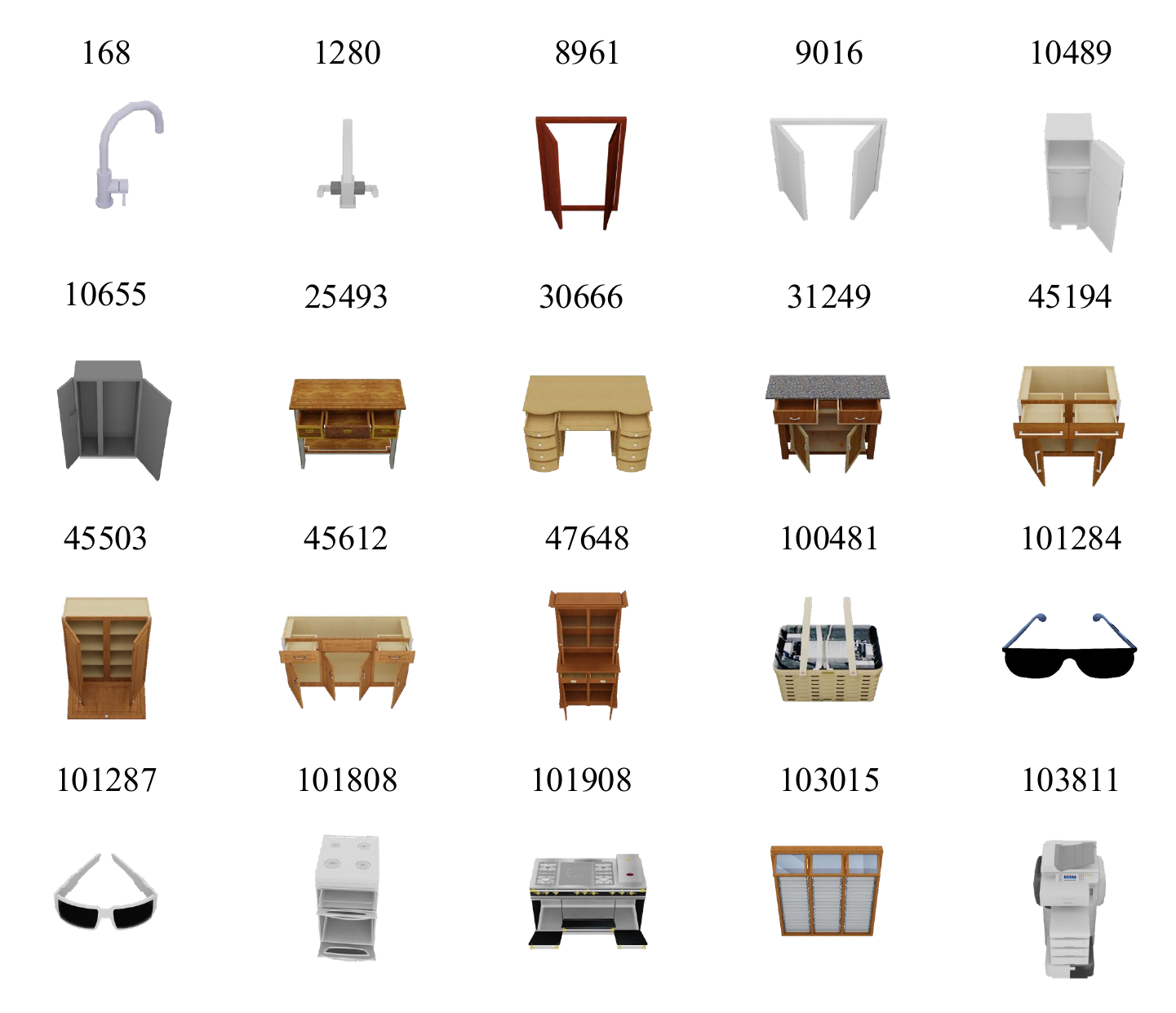}}
    \vspace{-10pt}
    \caption{{Visualization of \model-20 dataset.}}
    \label{fig:stats}
    \vspace{-10pt}
\end{figure}

We introduce and evaluate our method on \model-20, a newly curated dataset featuring 10 object categories: Faucet, Door, Refrigerator, Table, Storage Furniture, Bucket, Eyeglasses, Oven, Window, and Printer. For each object, we render a monocular video with 150 static frames in different viewpoints and 60 dynamic frames for each movable part.
The dataset provides a challenging benchmark with objects containing up to 10 parts and 9 movable joints. Further details and visualizations are available in~\cref{tab:app:data_conf} and~\cref{fig:stats}.

\subsection{3D Gaussian Splatting}\label{app:3dgs}
\ac{3dgs} represents a 3D scene using a collection of 3D Gaussians~\citep{kerbl20233d}. Each Gaussian $G_i$ is parameterized by its center $\vmu_i \in \mathbb{R}^3$, covariance matrix $\mSigma_i \in \mathbb{R}^{3 \times 3}$, opacity $\sigma_i \in [0,1]$, and spherical harmonics coefficients $\vh_i$ for view-dependent color. The opacity of a 3D Gaussian at spatial point $\vx$ is computed as:
\begin{equation}
\begin{aligned}
\alpha_i(\vx)= \sigma_i \exp\left(-\frac{1}{2}(\vx-\vmu_i)^\top\mSigma_i^{-1}(\vx - \vmu_i)\right), && \text{where} && \mSigma_i = \mR_i\mS_i\mS_i^\top\mR_i^\top.
\end{aligned}
\label{eq:gs_opacity}
\end{equation}
To ensure $\mSigma_i$ remains positive semi-definite, it is decomposed into a rotation matrix $\mR_i$ (parameterized by quaternion $\vr_i$) and a scaling diagonal matrix $\mS_i$ (parameterized by scale vector $\bm{s}_i$). 
To render an image, 3D Gaussians are projected onto the 2D image plane and aggregated using $\alpha$-blending:
\begin{equation}
\begin{aligned}
\mI = \sum_{i=1}^{N}T_i\alpha_i^{\text{2D}}\mathcal{SH}(\vh_i, \vv_i), && \text{where} && T_i = \prod_{j=1}^{i-1}(1 - \alpha_j^{\text{2D}}).
\end{aligned}
\label{eq:gs_rendering}
\end{equation}
Here, $\alpha_i^{\text{2D}}$ is the 2D version of~\cref{eq:gs_opacity}, $\mathcal{SH}(\cdot)$ calculates spherical harmonics for view direction $\vv_i$. Given multi-view images $\{\bar{\mI}_i\}_{i=1}^{N}$, \ac{3dgs} optimizes the paramters using L1 loss and D-SSIM loss~\citep{kerbl20233d} with a loss weight $\lambda_{\text{SSIM}}$:
\begin{equation}
\begin{aligned}
\mathcal{L}_{I} = (1-\lambda_{\text{SSIM}})\mathcal{L}_1 + \lambda_{\text{SSIM}}\mathcal{L}_{\text{D-SSIM}},
\end{aligned}
\label{eq:rendering_loss}
\end{equation}

\begin{table*}[t]
\centering
\caption{Dataset configuration.}
\label{tab:app:data_conf}
\renewcommand{\arraystretch}{1.2}
\begin{tabular}{cccccc}
\toprule
Object ID & Category         & \#Part & \#Joint & \#Revolute & \#Prismatic \\
\midrule
168       & Faucet           & 3      & 2       & 2          & 0           \\
1280       & Faucet           & 3      & 2       & 2          & 0           \\
8961      & Door             & 3      & 2       & 2          & 0           \\
9016      & Door             & 3      & 2       & 2          & 0           \\
10489     & Refrigerator     & 3      & 2       & 2          & 0           \\
10655     & Refrigerator     & 3      & 2       & 2          & 0           \\
25493     & Table            & 4      & 3       & 0          & 3           \\
30666     & Table            & 10     & 9       & 0          & 9           \\
31249     & Table            & 5      & 4       & 2          & 2           \\
45194     & Storage Furniture & 5      & 4       & 2          & 2           \\
45503     & Storage Furniture & 4      & 3       & 3          & 0           \\
45612     & Storage Furniture & 7      & 6       & 4          & 2           \\
47648     & Storage Furniture & 7      & 6       & 4          & 2           \\
100481    & Bucket           & 3      & 2       & 2          & 0           \\
101284    & Eyeglasses       & 3      & 2       & 2          & 0           \\
101287    & Eyeglasses       & 3      & 2       & 2          & 0           \\
101808    & Oven             & 3      & 2       & 2          & 0           \\
101908    & Oven             & 4      & 3       & 3          & 0           \\
103015    & Window           & 4      & 3       & 3          & 0           \\
103811    & Printer          & 7      & 6       & 0          & 6           \\
\midrule
Average   & ---              & 4.35   & 3.35    & 2.05       & 1.3        \\
\bottomrule
\end{tabular}
\end{table*}

\subsection{Articulation Modeling}
\label{app:imp:arti}
\paragraph{Articulation Modeling}
Building upon the part assignments, we model articulation through learnable joint parameters, including axis direction $\vd$, axis origin $\vo$, and time-variant joint state $\theta^t$. To learn a smooth trajectory of joint states, we model it with Foriour embedding $E(\cdot)$ followed by a learnable MLP:
$\theta^t=\text{MLP}(E(t))$. We represent the rigid transformation as dual-quarternion $\vq^t=(\vq_{r}^t,\vq_{d}^t)$ for smooth skinning, where $\vq_{r}^t,\vq_{p}^t$ represent the rotation and translation components respectively. 
The dual-quaternion of each joint could be calculated as:
\begin{equation}
\begin{aligned}
& \text{prismatic}: \vq_{r}^t = (1, 0, 0, 0), \quad
\bar{\vo}^t=(0, \theta^t \cdot \vd), \quad
\vq_d^t=0.5 \cdot \bar{\vo}^t \otimes \vq_{r}^t,  \\
& \text{revolute} : \vq_{r}^t = (\cos\frac{\theta^t}{2}, \sin\frac{\theta^t}{2}\cdot \vd), \quad
\bar{\vo}^t=(0, \vo), \quad
\vq_{d}^t = 0.5 \cdot (\bar{\vo}^t \otimes \vq_{r}^t -  \vq_{r}^t \otimes \bar{\vo}^t). \\
\end{aligned}
\end{equation}
Then we calculate the per-gaussian dual-quaternion $\vq_i^t$ with part assignment probabilities $\vm_i$ by:
\begin{equation}
\begin{aligned}
\vq_i^t =\sum_{k=1}^K m_{ik}\cdot {\vq^t_{k}}=(\sum_{k=1}^K m_{ik}\cdot {\vq^t_{k,r}},\sum_{k=1}^K m_{ik}\cdot {\vq^t_{k,d}}). 
\end{aligned}
\end{equation}
where $\vq_k^t$ is the dual-quaternion of $k$-th part. The position and rotation of Gaussian $G_i^t$ are obtained by:
\begin{equation}
\begin{aligned}
\vmu_i^t = \mR_i^{t}\cdot \vmu_i^c + \vt_i^{t}, \quad \vr_i^t = {\vq}_{i,r}^{t} \otimes \vr_i^c,\\
\end{aligned}
\label{eq:articulation_modeling}
\end{equation}
where $\mR_i^{t}$ and $\vt_i^{t}$ is rotation matrix and translation vector derived from $\vq_i^{t}$, and $\otimes$ denotes quaternion multiplication operation. 
We provide the detailed derivation process of dual-quaternion in the following paragraphs.

\paragraph{Dual Quaternions for SE(3) Transformation}
A general rigid SE(3) transformation in 3D space consists of a rotation followed by a translation. A dual quaternion represents this combined operation within a single algebraic entity.
Let the rotation be represented by a unit quaternion $\vq_r$ and the translation by a vector $\vt$. A point $\vp$ in space, represented as a pure quaternion $\bar{\vp} = (0, \vp)$, is transformed to a new point $\vp'$ by first applying the rotation and then the translation: $\vp' = \vq_r \otimes \vp \otimes \vq_r^* + \vt$, where $\vq_r^*$ is the conjugate of $\vq_r$ and $\otimes$ denotes the quaternion multiplication operation.

A dual quaternion $\vq$ is defined as $\vq = \vq_r + \varepsilon \vq_d$, where $\vq_r$ is the real part, $\vq_d$ is the dual part, and $\varepsilon$ is the dual unit with the property $\varepsilon^2 = 0$. Given the rotation quaternion $\vq_r$ and translation $\vt$, the dual part $\vq_d$ could be calculated by: $\vq_d = \frac{1}{2} (0, \vt) \otimes \vq_r$.

\paragraph{Dual Quaternions for Articulated Transformation}
We apply the above principles to derive the specific formulas for prismatic and revolute joints at time $t$.

\textbf{\textit{Prismatic}}: A prismatic joint executes a pure translation with no rotation, so that the real part is the unit quaternion $\vq_r^t = (1, 0, 0, 0)$. Given the axis direction $\vd$ and joint state $\theta^t$, its translation component is $\vt=\theta^t\cdot\vd$. Let $\bar{\vo}^t=(0, \theta^t\cdot\vd)$, the dual part could be calculated by: $\vq_d = \frac{1}{2} \bar{\vo} \otimes \vq_r^t$.

\textbf{\textit{Revolute}}: A revolute joint executes a pure rotation, not about the world origin, but about the joint's origin point $\vo$. Given the axis direction $\vd$, axis origin $\vo$ and joint state $\theta^t$ this "off-center" rotation is equivalent to a sequence of three operations: (1) translate the system so the pivot point $\vo$ moves to the origin: $\bar{\vo}^t=(0,\vo),\ \vq_{T_1} = 1 - \frac{\varepsilon}{2}\bar{\vo}$); (2) perform the rotation around the origin: $\vq_R=\vq_r^t = (\cos\frac{\theta^t}{2}, \sin\frac{\theta^t}{2}\cdot \vd)$. (3) translate the system back: $\vq_{T_2} = 1 + \frac{\varepsilon}{2}\bar{\vo}$. The total transformation $\vq^t$ is the product:
\begin{equation*}
\vq^t = \vq_{T_2}\vq_{R}\vq_{T_1} = (1 + \frac{\varepsilon}{2}\bar{\vo})\vq_r^t (1 - \frac{\varepsilon}{2}\bar{\vo})=(1 + \frac{\varepsilon}{2}\bar{\vo}) (\vq_r^t - \frac{\varepsilon}{2}\vq_r^t \bar{\vo})=\vq_r^t + \varepsilon \left( \frac{1}{2}\bar{\vo}\vq_r^t - \frac{1}{2}\vq_r^t\bar{\vo} \right), \\
\end{equation*}
where $\otimes$ is omitted for brevity.
As a result, the real part and dual part are calculated as: $\vq_r^t = (\cos\frac{\theta^t}{2}, \sin\frac{\theta^t}{2}\cdot \vd),\ \vq_d^t = \frac{1}{2} (\bar{\vo} \otimes \vq_r^t - \vq_r^t \otimes \bar{\vo})$.

\subsection{Initialzation and Optimization}
\paragraph{Inversed Deformation Field}\label{app:inv_df}
Given a point position $\vx_i^{t_0}$ sampled from the trajectory $\{\vx^t_i\}_{t=1}^{T}$, we extend our part assignment module from canonical space to observation space to obtain the part assignment probabilities $\vm_i^{t_0}$ of $\vx^{t_0}_i$ at time $t_0$. Specifically, we deform the learnable centers $C_k=(\vp_k,\mV_k, \lambda_k)$ from canonical space to observation space by:
\begin{equation}
\begin{aligned}
\vp_k^{t_0} = \mR_k^{t_0}\cdot \vp_k + \vt_k^{t_0}, \quad \mV_k^{t_0} = \mR_k^{t_0}\cdot \mV_k, \quad C_k^{t_0}=(\vp_k^{t_0}, \mV_k^{t_0}, \lambda_k),\\
\end{aligned}
\end{equation}
where $\mR_k^{t_0}$ and $\vt_k^{t_0}$ is rotation matrix and translation vector derived from $\vq_k^{t_0}$. We replacing $C_k$ with $C^{t_0}_k$ in \cref{eq:dist} and \cref{eq:part_assignment} to calculate $\vm^{t_0}_i$, then the canonical position $\hat \vx^c_i$ is calculated by: 
\begin{equation}
\begin{aligned}
\vq_i^{t_0}=\sum_{k=1}^Km_{ik}^{t_0}\cdot \vq_k^{t_0}, \quad \hat\vx_i^{c} = (\mR_i^{t_0})^{-1}\cdot (\vx_i^{t_0} - \vt_i^{t_0}),\\
\end{aligned}
\end{equation}
where $\mR_i^{t_0}$ and $\vt_i^{t_0}$ is rotation matrix and translation vector derived from $\vq_i^{t_0}$.

\paragraph{Training Configuration} We train deformation field $\mathcal{F}$ for 10K steps with loss $\mathcal{L}_\text{track}=\mathcal{L}_{o2o}+\mathcal{L}_{c2o}$ described in~\cref{eq:track_c2o} and ~\cref{eq:track_o2o}, taking 5-10 minutes per object.
We train canonical Gaussians $\mathcal{G}^c$ for 20K steps with loss $\mathcal{L}_\text{render}=(1-\lambda_\text{{SSIM}})\mathcal{L}_1+\lambda_\text{{SSIM}}\mathcal{L}_{\text{D-SSIM}}+\mathcal{L}_D$, where $\lambda_\text{{SSIM}}=0.2$ is used in experiments. This stage takes about 4 minutes per object.
We jointly optimize the canonical Gaussians and deformation field for 20K steps with $\mathcal{L}=\mathcal{L}_I+\mathcal{L}_D+\lambda_{c2o}\mathcal{L}_{c2o}$, where $\lambda_{c2o}=0.5$. This state takes 10-20 minutes per object. 

\subsection{Joint Type Prediction Using GPT-4o}
Inspired by SINGAPO~\citep{liu2024singapo}, we use GPT-4o to predict the number of joints and joint types. We input the video and a step-by-step instruction to make GPT-4o understand the articulated objects. The version of GPT-4o used in our experiments is gpt-4o-2024-11-20. The instruction is:
\begin{figure}[htbp]
    \centering
    \includegraphics[width=\textwidth]{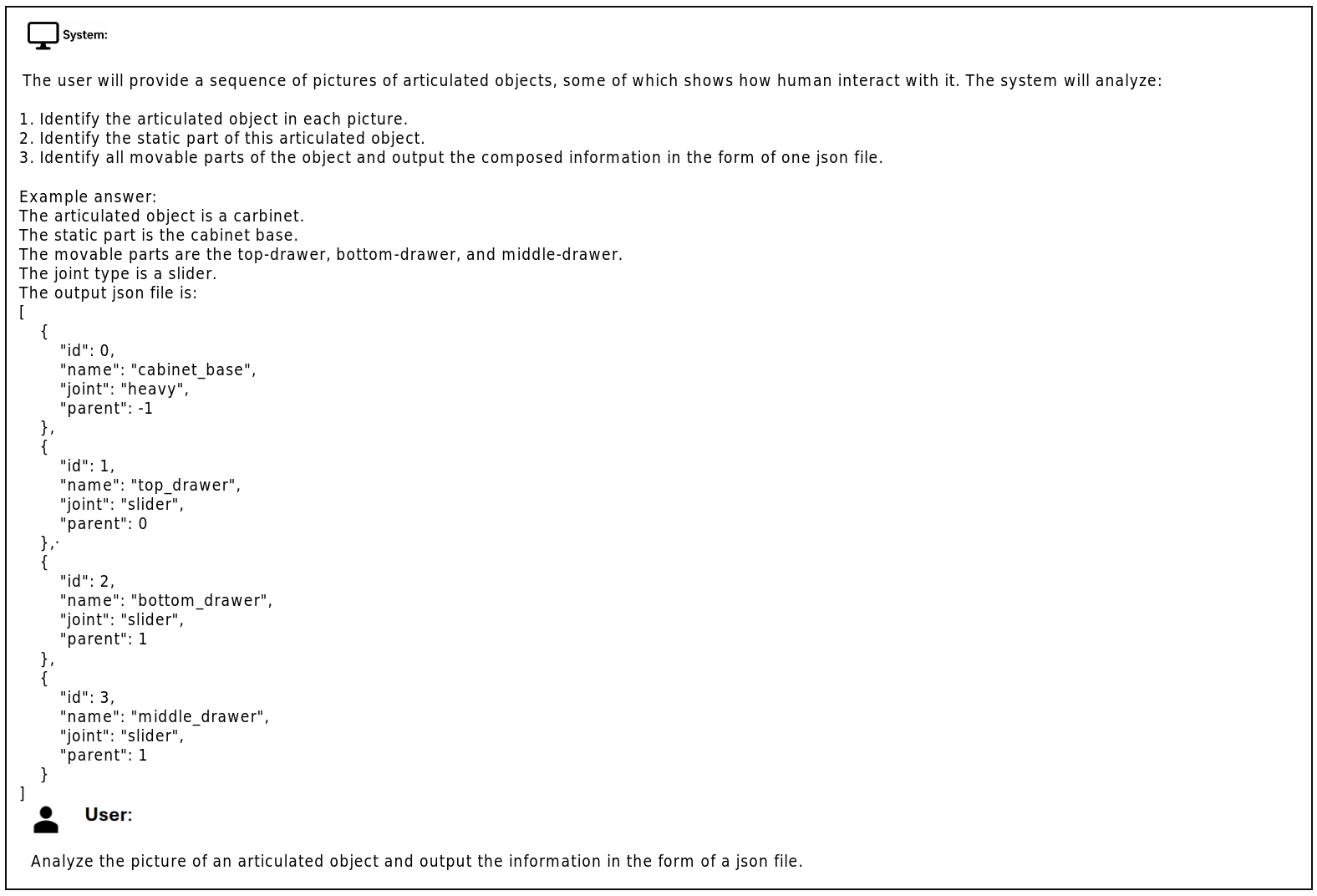}
    \caption{Prompt for GPT-4o to predict the number of joints and joint types.}
    \label{fig:gpt4o_prompt}
\end{figure}

\section{Limitations}
\label{app:limitation}
Our method, while effective, has limitations that open avenues for future research.
\paragraph{Dependency on Upstream Perception Models.} The final quality of our reconstruction is inherently dependent on the accuracy of the upstream models used for perception. Our pipeline first relies on a monocular depth and camera pose estimator (e.g., VGGT). Subsequently, a pre-trained tracking model (e.g., TAPIP3D) generates 3D motion tracks. If the depth or camera pose estimates contain significant errors, the resulting 3D tracks will be noisy and fail to capture the object's true rigid-body motion. This can lead to failures in our downstream fitting and clustering steps, resulting in distorted geometry or incorrect joint estimation. However, as this is a rapidly advancing field, we anticipate that progress in visual foundation models and tracking models will continue to mitigate this dependency.

\paragraph{Canonical Gaussian Initialization.} Our current framework assumes that the input video begins with a short sequence (N frames) where the scene is static. This segment is crucial for initializing the canonical Gaussian representation of the object's geometry. While this assumption is practical for data captured by a user (self-shot), it restricts the method's applicability to in-the-wild videos from the internet, which often begin with immediate motion. Relaxing this condition is non-trivial, as it makes the ill-posed problem of disentangling geometry from motion even more challenging. A promising direction for future work is to incorporate powerful generative priors. Such models could help infer a plausible canonical shape even from a video with continuous motion, thereby enabling reconstruction from arbitrary monocular inputs.

\paragraph{Reliance on Motion for Part Segmentation.} Our approach infers part segmentation exclusively from motion cues by clustering the derived 3D tracks. This reliance on dynamics places high demands on the tracking quality and can be fragile for objects with many parts or for parts that exhibit very subtle relative motion. In such challenging cases, the segmentation quality can degrade, leading to incorrectly merged or split components. A valuable future direction is to augment our motion-based clustering with appearance-based priors from pre-trained foundation models. For instance, integrating semantic features from DINOv2~\citep{oquab2023dinov2} or segmentation masks from models like SAM~\citep{kirillov2023segment} could provide a powerful, independent signal for identifying object parts, making the segmentation process significantly more robust.


{\subsection{Additional Qualitative Results}}
{We provide additional qualitative results in the following pages.}
\begin{figure}[t!]
 \centering
 \resizebox{0.9\linewidth}{!}{\includegraphics[width=\linewidth]{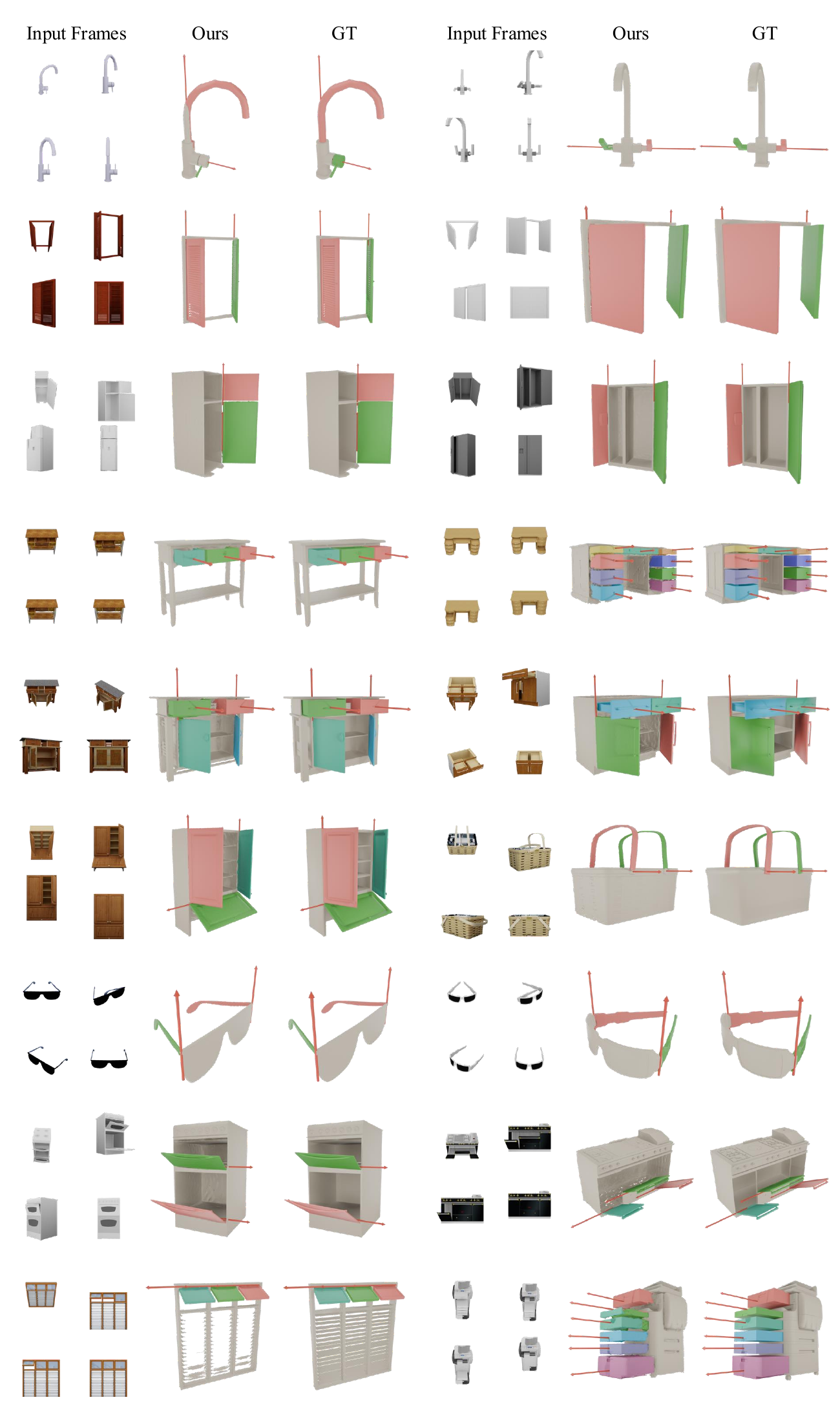}}
 \caption{\textbf{Additional qualitative results on \model-20.}}
 \label{fig:app:mpart0}
\end{figure}

\end{document}